\documentclass[10pt,twocolumn,letterpaper]{article}

\usepackage{iccv}
\usepackage{times}
\usepackage{epsfig}
\usepackage{graphicx}
\usepackage[numbers,sort&compress]{natbib}
\usepackage{amsmath}
\usepackage{amssymb}
\usepackage{booktabs}

\usepackage{algorithm,algpseudocode}
\algnewcommand{\algorithmicforeach}{\textbf{for each}}
\algdef{SE}[FOR]{ForEach}{EndForEach}[1]
  {\algorithmicforeach\ #1\ \algorithmicdo}
  {\algorithmicend\ \algorithmicforeach}

\usepackage{pifont}

\usepackage{color, colortbl}
\definecolor{Gray}{gray}{0.93}
\definecolor{Orange}{rgb}{1,0.5,0}
\definecolor{DGray}{gray}{0.83}
\definecolor{LightCyan}{rgb}{0.88,1,1}

\usepackage{wrapfig}

\usepackage[dvipsnames]{xcolor}

\usepackage{multirow,mathtools } \usepackage{algorithm,algpseudocode}

\usepackage{pifont}
\usepackage{color, colortbl}

\usepackage{blindtext}
\usepackage{lipsum}

\usepackage{multirow}
\usepackage{graphicx}
\usepackage{listings}

\usepackage[most]{tcolorbox}

\usepackage{amsmath,amsfonts,bm}

\def\eqref#1{(\ref{#1})}

\def\1{\bm{1}}

\DeclareMathAlphabet{\mathsfit}{\encodingdefault}{\sfdefault}{m}{sl}
\SetMathAlphabet{\mathsfit}{bold}{\encodingdefault}{\sfdefault}{bx}{n}

\DeclareMathOperator*{\argmin}{arg\,min}

\DeclareMathOperator*{\minimize}{\text{minimize}}

\DeclareMathOperator*{\st}{\text{subject to}}

\newcommand{\btheta}{\boldsymbol{\theta}}

\newcommand{\bdelta}{\boldsymbol\delta}

\newcommand{\bx}{\mathbf{x}}

\newcommand{\bphi}{\boldsymbol\phi}
\newcommand{\bpsi}{\boldsymbol\psi}
\usepackage[pagebackref,breaklinks,colorlinks]{hyperref}
\usepackage[capitalize]{cleveref}
\usepackage{bbding}

\newcommand{\ours}{\textsc{AdvMoE}}

\newcommand{\trades}{\textsc{TRADES}}
\newcommand{\AdvRT}{\textsc{AT}}

\newcommand{\denseRT}{{AT (Dense)}}
\newcommand{\slimRT}{{AT (S-Dense)}}
\newcommand{\sparseRT}{{AT (Sparse)}}
\newcommand{\moeRT}{AT (MoE)}

\newcommand{\dense}{S-Dense}
\newcommand{\ldense}{Dense}

\newcommand{\pcnn}{Sparse-CNN}
\newcommand{\moe}{MoE-CNN}

\newcommand{\cifarten}{\texttt{CIFAR-10}}
\newcommand{\cifarhun}{\texttt{CIFAR-100}}
\newcommand{\imagenet}{\texttt{ImageNet}}
\newcommand{\timagenet}{\texttt{TinyImageNet}}

\makeatletter
\def\blfootnote{\gdef\@thefnmark{}\@footnotetext}
\makeatother

\definecolor{my_purple}{HTML}{c1ace8}

\definecolor{lightorange}{HTML}{fc8e62}
\definecolor{lightgray}{gray}{0.6}

\newcommand{\orangecolor}[1]{\textcolor{lightorange}{#1}}
\newcommand{\partcirc}{\orangecolor{$\mathbf{\circ}$\,}}
\newcommand{\fullcirc}{\orangecolor{$\bullet$\,}}
\newcommand{\gr}{\rowcolor[gray]{.95}}

\iccvfinalcopy 

\def\iccvPaperID{5132}

\begin{document}

\title{Robust Mixture-of-Expert Training for Convolutional Neural Networks}

\author{
Yihua Zhang\textsuperscript{1}
\and
Ruisi Cai\textsuperscript{2}
\and
Tianlong Chen\textsuperscript{2,3,4,5}
\and
Guanhua Zhang\textsuperscript{6}
\and
Huan Zhang\textsuperscript{7,8}
\and
Pin-Yu Chen\textsuperscript{9}
\and
Shiyu Chang\textsuperscript{6}
\and
Zhangyang Wang\textsuperscript{2}
\and
Sijia Liu\textsuperscript{1,9}
\and
\textsuperscript{1}Michigan State University,
\textsuperscript{2}University of Texas at Austin,\\
\textsuperscript{3}The University of North Carolina at Chapel Hill, \textsuperscript{4} MIT, \textsuperscript{5}Harvard University,\\
\textsuperscript{6}UC Santa Barbara, 
\textsuperscript{7}Carnegie Mellon University,
\textsuperscript{8}UIUC,
\textsuperscript{9}IBM Research\\
}

\maketitle
\blfootnote{Correspondence to: Yihua Zhang$<$\href{mailto:zhan1908@msu.edu}{zhan1908@msu.edu}$>$} 
\ificcvfinal\thispagestyle{empty}\fi

\begin{abstract}
Sparsely-gated Mixture of Expert ({MoE}), an emerging  deep model architecture,  has  demonstrated a great promise to enable high-accuracy and ultra-efficient model inference. 
Despite the growing popularity of MoE, little work investigated its potential to advance convolutional neural networks (CNNs), especially in the plane of \textit{adversarial robustness}. Since the lack of robustness has become one of the main hurdles for CNNs, in this paper we ask: How to adversarially robustify a CNN-based MoE model?
Can we robustly train it like an ordinary CNN model? 
Our pilot study shows that the conventional adversarial training (AT) mechanism (developed for vanilla CNNs) no longer remains effective to robustify an MoE-CNN.
To better understand this phenomenon, we dissect the robustness of an MoE-CNN into two dimensions: Robustness of routers (\textit{i.e.}, gating functions to select data-specific experts) and   robustness of experts (\textit{i.e.}, the router-guided pathways defined by the subnetworks of the backbone CNN). Our analyses show that   routers and experts are hard to adapt to each other in the vanilla AT. 
Thus, we propose a new router-expert alternating \underline{Adv}ersarial training framework for \underline{MoE}, termed {\ours}.
The effectiveness of our proposal is justified across 4 commonly-used CNN model architectures over 4 benchmark datasets. We find that {\ours} achieves $1\% \sim 4\%$  adversarial robustness improvement over the original dense CNN, and enjoys the efficiency merit of sparsity-gated MoE, leading to more than $50\%$ inference cost reduction. Codes are available at \url{https://github.com/OPTML-Group/Robust-MoE-CNN}.
\end{abstract}

\section{Introduction}
\label{sec: intro}
\begin{figure*}[t]
    \centering
       \begin{tabular}{cc}
       \hspace*{-3mm}\includegraphics[width=.5\textwidth,height=!]{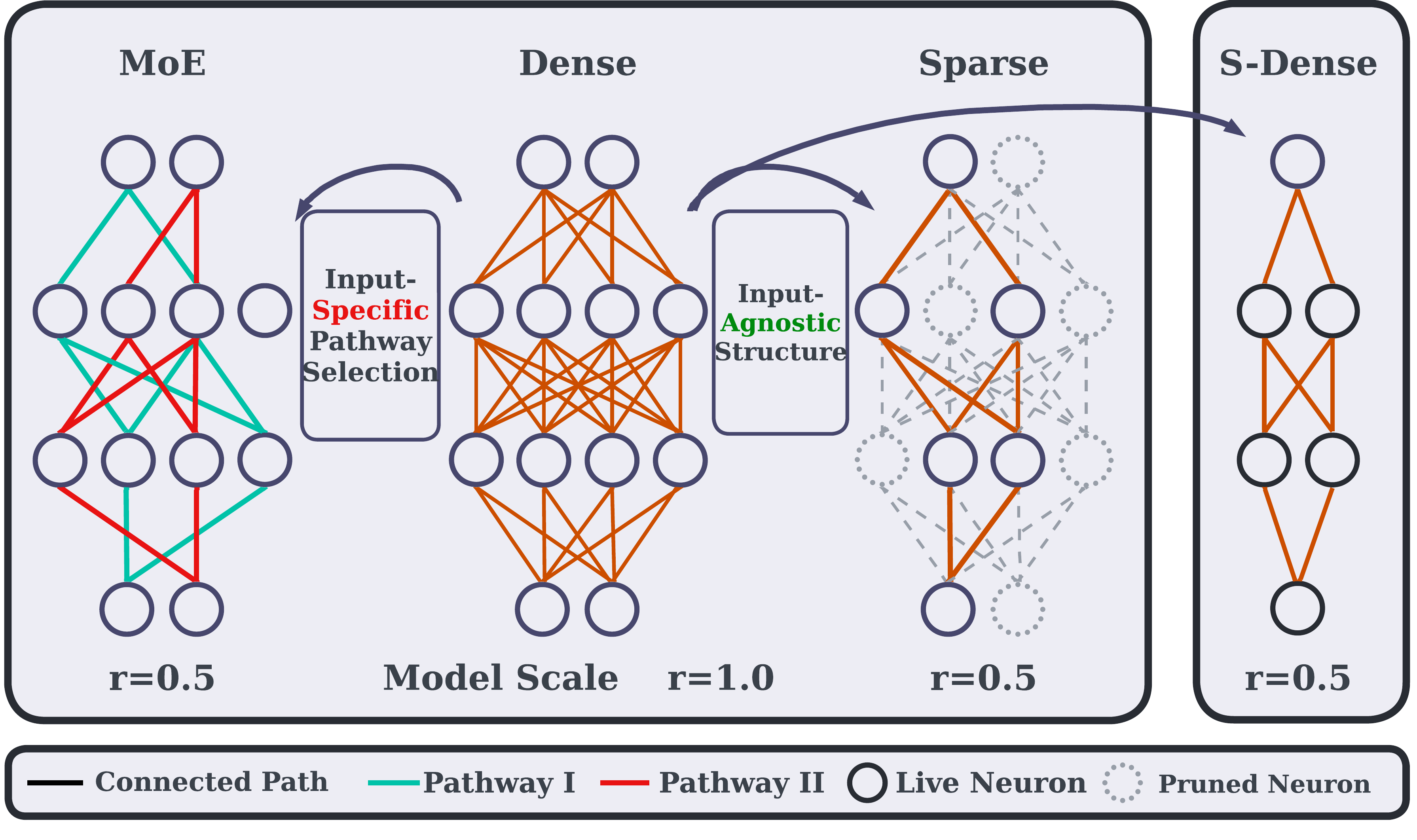}  &
       \hspace*{-3mm}\includegraphics[width=.5\textwidth,height=!]{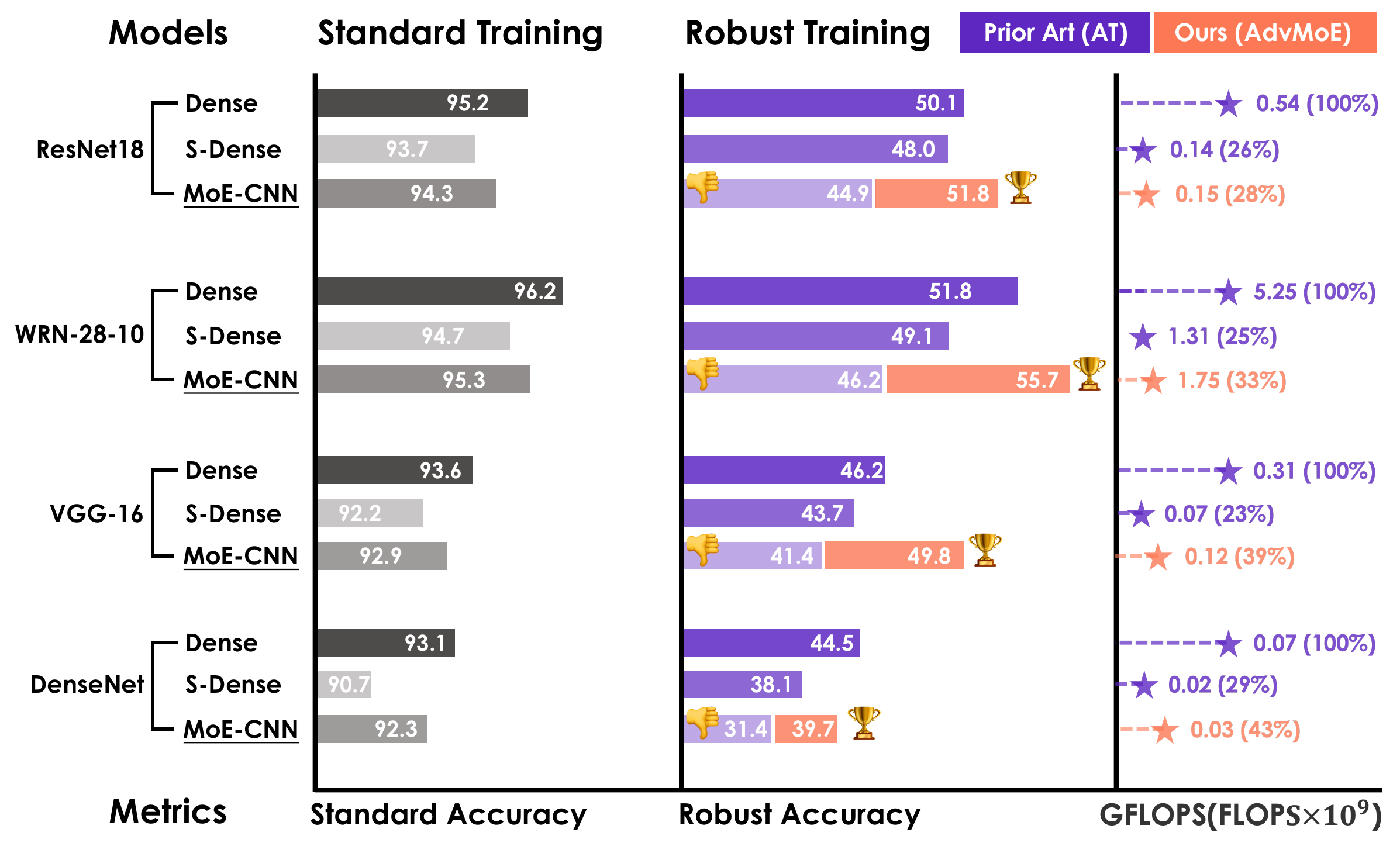} \vspace*{-2mm}\\
        \footnotesize{(a) Illustration of  CNN types considered   in this work.} & \footnotesize{(b) Performance overview on CIFAR-10.}
        \end{tabular}
    \caption{{(a) Model types ({\ldense}, {\moe}, {\pcnn}, and {S(mall)-Dense}) considered in this paper; see details in `Model setup' of Sec.\,\ref{sec: prob_state}.
    (b) Performance overview using the standard training and the robust training on model architectures in (a), where standard accuracy and robust accuracy are defined by testing accuracy on the benign and adversarial test datasets, respectively.   
    Compared to standard training (results in  \textcolor{gray}{\textbf{gray}}), 
    the conventional  AT \cite{zhang2019theoretically}   is no longer effective for {\moe}  (see   results in \textcolor{my_purple}{\textbf{light purple}}). This is in contrast to AT for other CNN models ({\ldense} and {\dense}). Different from AT, 
     our proposed {\ours} can effectively equip {\moe} with   improved robustness, higher than {\ldense} (see results in \textcolor{orange}{\textbf{orange}}), without losing its   inference efficiency (see ``GFLOPS''). We refer readers to Sec.\,\ref{sec: exp_setup} for more experiment details.
    }}
    \label{fig: overview}
\end{figure*}

Despite the state-of-the-art performance achieved by the outrageously large networks\,\cite{arnab2021vivit, dosovitskiy2020image, kolesnikov2020big, raffel2020exploring, sun2017revisiting} in various deep learning (DL) tasks, it still remains challenging to train and deploy such models cheaply. A major bottleneck is the lack of parameter efficiency\,\cite{zhang2021moefication}: A single data   prediction only  requires activating a small portion of   the parameters of the full model.
Towards efficient DL, sparse Mixture of Experts (MoE)\,\cite{shazeer2017outrageously, wang2020deep, riquelme2021scaling, fedus2021switch, xue2022go, komatsuzaki2022sparse, zhou2022mixture, li2022sparse, chen2023sparse} aims to divide and conquer the model parameters based on their optimal responses to  specific inputs so that  inference costs can be reduced. 
A typical MoE structure is comprised of a set of `experts' (\textit{i.e.}, sub-models extracted from the original backbone network)  and  `routers' (\textit{i.e.}, additional small-scale gating networks to determine expert selection schemes across layers). 
During inference,  sparse MoE only activates the most relevant  experts and forms the expert-guided pathway for a given  input data. By doing so, sparse MoE can boost the inference efficiency (see `GFLOPS' measurement in \textbf{Fig.\,\ref{fig: overview}}).
Architecture-wise, sparse MoE has been used for both CNNs \cite{wang2020deep, gross2017hard} and vision transformers (ViTs) \cite{shazeer2017outrageously, riquelme2021scaling, fedus2021switch, xue2022go, komatsuzaki2022sparse, zhou2022mixture, li2022sparse, chen2023sparse, liu2021swin}. Yet, we will   focus on the former since sparse MoE for CNNs is under-explored compared to   non-sparse MoE  for CNNs \cite{abbas2020biased,ahmed2016network,pavlitskaya2020using}, and adversarial robustness (another key performance metric of our work) was extensively studied in the context of CNNs.

It is known that a main weakness of DL is the    lack of adversarial robustness    \cite{goodfellow2014explaining,carlini2017towards,papernot2016limitations}. For example, CNNs can be easily fooled by adversarial attacks\,\cite{goodfellow2014explaining,carlini2017towards,papernot2016limitations}, in terms of tiny input perturbations generated to direct to  erroneous predictions. Thus, adversarial training (\textbf{AT}) of CNNs has become a main research thrust  \cite{madry2018towards,zhang2019theoretically,shafahi2019adversarial,wong2020fast,zhang2019you,athalye2018obfuscated}. However, when CNN meets   sparse MoE, it remains elusive if the improved inference efficiency brought by the sparse MoE   comes at the cost of more complex adversarial training recipes.
Thus, we ask:

\vspace{0.1em}
\begin{tcolorbox}[before skip=0.2cm, after skip=0.2cm, boxsep=0.0cm, middle=0.1cm, top=0.1cm, bottom=0.1cm]
\textit{\textbf{(Q)} What will be the new insights into   adversarial robustness of   sparse MoE-integrated CNNs? And   what will be the suited AT mechanism?}
\end{tcolorbox}
\vspace{0.1em}

To our best knowledge, problem \textbf{(Q)} remains open in the literature.
The most relevant work to ours is \cite{puigcerver2022adversarial}, which investigated the adversarial robustness   of MoE and leveraged the ordinary AT recipe \cite{madry2018towards} to defend against adversarial attacks. However, it only focused  on the ViT architecture, making a vacancy for the research on robustification for the sparse MoE-based CNN (termed \textbf{{\moe}} in this work). Most importantly, we find that the vanilla AT \cite{madry2018towards, zhang2019theoretically} (widely used to robustify CNNs) is \textit{no longer} effective for {\moe}.
Thus, new solutions are in demand. 

To address \textbf{(Q)}, we need to (1) make careful sanity checks for  AT in {\moe}, (2) make an in-depth analysis of its  failure cases, and (3) advance new AT principles that can effectively improve robustness without losing the generalization and efficiency from sparse MoE. Specifically, our \textbf{contributions} are unfolded below.
\begin{itemize}
    \item  We dissect the MoE robustness into two new dimensions (different from CNNs): routers' robustness and experts' robustness. Such a robustness dissection brings novel insights into the (in)effectiveness   of  AT.
    \item Taking inspiration from   the above robustness dissection,  we propose a new \underline{Adv}ersarial training framework for \underline{MoE}, termed {\ours}, which enforces routers and experts to make a concerted effort to improve the overall robustness of {\moe}.
    \item We conduct extensive experiments to demonstrate the effectiveness of {\ours} across $4$ CNN architectures and 4 datasets. For example, {\ours} outperforms AT on the original dense CNN model (termed {\ldense}) by a substantial margin:  $1\%\sim 4\%$ adversarial robustness improvement and over $50\%$ reduction of  inference overhead; see \textbf{Fig.\,\ref{fig: overview}} for illustrations on different CNN types  and highlighted performance achieved.
\end{itemize}

\section{Related Work}
\label{sec: related_work}
\paragraph{Sparsely-activated Mixture of Experts (Sparse MoE).} As a special instance of compostional neural architectures \cite{yang2022deep,liu2022dataset,yang2022KF}, MoE~\cite{raffel2020exploring, shazeer2017outrageously, wang2020deep, riquelme2021scaling, fedus2021switch, xue2022go,gross2017hard, abbas2020biased,ahmed2016network, pavlitskaya2020using, yuksel2012twenty, jordan1994hierarchical, eigen2013learning, jacobs1991adaptive, chen1999improved, lewis2021base, lepikhin2020gshard, du2022glam} aims at solving ML tasks in a divide-and-conquer fashion, which creates a series of sub-models (known as the \textit{experts}) and conducts input-dependent predictions by combing the output of sub-models. As an important branch of MoE, sparsely gated MoE \cite{shazeer2017outrageously, wang2020deep, riquelme2021scaling, fedus2021switch, xue2022go, komatsuzaki2022sparse, zhou2022mixture, li2022sparse, chen2023sparse, lepikhin2020gshard, zoph2022designing, mustafa2022multimodal, du2022glam, lewis2021base, raffel2020exploring, gross2017hard} only activates a subset of experts based on a routing system. The major advantage brought by sparse MoEs lies in its sub-linear increasing inference costs (FLOPs) with respect to (\textit{w.r.t.}) model scales (parameter counts) \cite{shazeer2017outrageously}. 
In the vision domain, a vast majority of the existing works focus on the design of MoE for ViTs \cite{lepikhin2020gshard, mustafa2022multimodal, riquelme2021scaling, fedus2021switch, xue2022go, komatsuzaki2022sparse, zhou2022mixture, li2022sparse, chen2023sparse, zoph2022designing}, leaving MoE for CNNs under-explored \cite{gross2017hard, wang2020deep}. To our best knowledge, DeepMoE\,\cite{wang2020deep} is the most recent work that systematically studies the integration of MoE with CNNs, but restricts to the standard (non-robust) training paradigm. Meanwhile, there also exist other works related to  {\moe}, but they either fall out of the ``sparse'' MoE scope \cite{abbas2020biased, pavlitskaya2020using} or bring no efficiency gains \cite{ahmed2016network}. By contrast, we focus on the \textit{efficiency-promoting} {\moe} setup throughout this work. 

\paragraph{Adversarial robustness.} CNNs are notoriously vulnerable to imperceptible adversarial samples~\cite{carlini2017towards,papernot2016limitations,szegedy2013intriguing} and thus training adversarially robust models\,\cite{goodfellow2014explaining,kurakin2016adversarial,madry2018towards} has become a main research focus in many areas. Most of the robust training methods \cite{madry2018towards,zhang2019theoretically,shafahi2019adversarial,wong2020fast,zhang2019you,athalye2018obfuscated} are extended from min-max optimization-based adversarial training\,\cite{madry2017towards}. For instance, the work \cite{zhang2019theoretically} seeks an optimal balance between robustness and standard generalization ability. Other work \cite{shafahi2019adversarial,zhang2020attacks,zhang2022advancing,gui2019model,sehwag2020hydra,fu2021drawing, andriushchenko2020understanding, zhang2019you, chen2021robust, rebuffi2021data} aims at trimming down the computational costs of robust training while maintaining robustness.
The work~\cite{puigcerver2022adversarial}  studies the robustness of MoE-based architectures for the first time. Yet, its focus stays on MoE for ViTs and the relationship between model capacity and robustness.

\section{Problem Statement}
\label{sec: prob_state}
In this section, we start by presenting the setup of {\moe} in this work and then introduce  the robust learning paradigm. The lack of adversarial robustness of deep models inspires us to investigate whether the adversarial training ({\AdvRT}) approach designed for vanilla CNNs keeps effective for {\moe}. 
Through a motivating example, we show that
the conventional {\AdvRT} recipe is \textit{incapable} of 
equipping {\moe} with desired robustness. The resulting performance is  even worse than that of {\AdvRT}-produced {\dense}, which has a much smaller model capacity than {\moe}.
Thus, the question of how to robustify {\moe} arises.

\paragraph{Model setup.}
We consider a CNN-backboned MoE that consists of multiple MoE layers. Each MoE layer involves a router and a vanilla convolutional layer from the backbone CNN model. Within one MoE layer, we define $N$ experts,
each of which picks a subset of the channels from the convolutional layer. Specifically, suppose the $l$-th layer contains $C_l$ channels, one expert will contain $r \times C_l$ channels, where we call the ratio $r \in [0,1]$  \textit{model scale} and keep it the same across different layers (see \textbf{Fig.\,\ref{fig: overview}a}).
It is worth noting that as $r$ increases, the per-expert model capacity  increases  (\textit{i.e.}, with more parameters)  at the cost of the efficiency reduction.
In a forward path, the router first makes an \textit{input-specific} expert selection. These selected layer-wise experts then form an end-to-end pathway to process this input. We use ``\textit{pathway}'' to describe one experts-guided forward path (see \textbf{Fig.\,\ref{fig: overview}a}). We summarize  the model setup in \textbf{Fig.\,\ref{fig: moe_structure}}. 

Further, we introduce different model types  considered in this work and shown in   \textbf{Fig.\,\ref{fig: overview}a}. First, we term the original dense CNN   model `\textbf{\ldense}', which serves as the \textit{model basis} for other model types that derive from. Second, we directly shrink the channel number of each layer in {\ldense} (based on the model scale parameter $r$) to obtain the  `\underline{s}mall \underline{dense}' model (termed `\textbf{\dense}'). Notably, {\dense} has the size \textit{equivalent to a single pathway} in {\moe}. Third, we   use the structured pruning method\,\cite{sehwag2020hydra} to create a sparse subnetwork from {\ldense}, with the weight remaining ratio   same as the model scale parameter $r$ in {\moe}, which we call `\textbf{\pcnn}'. In summary,  {\dense} has the smallest model capacity (comparable to a single pathway of {\moe}), and   should provide the \textit{performance lower-bound} for {\moe}. By contrast, {\pcnn} has a larger model capacity but is smaller than {\moe} as it encodes a data-agnostic pathway of {\ldense}, while {\moe} yields data-specific pathways at the same scale. {\ldense} has the largest model capacity but the least inference efficiency.

\vspace*{2mm}
\noindent\textbf{Adversarial robustness: From CNN to MoE-CNN.}
It has been known that current machine learning models (\textit{e.g.}, CNNs) are vulnerable to adversarial attacks\,\cite{goodfellow2014explaining,carlini2017towards,papernot2016limitations}. Towards the robust design, a variety of  {\AdvRT} (adversarial training) methods have been developed. The predominant ones include  the min-max optimization-based vanilla {\AdvRT} \,\cite{madry2018towards} and its {\trades} variant \,\cite{zhang2019theoretically}  that strikes a balance between   generalization and adversarial robustness. Throughout the paper, we adopt {\trades} as the default conventional {\AdvRT}  recipe, which solves the following    problem: 

\vspace*{-2mm}
{\small{
\begin{align}
\displaystyle
\raisetag{2em}
\text{\resizebox{7.32cm}{!}{$
\hspace*{-2mm}\min_{\btheta} \mathbb{E}_{(\bx, y) \in \mathcal{D}} \left [\ell(\btheta; \bx, y) + \frac{1}{\lambda} \max_{\| \bdelta \|_\infty \leq \epsilon } \ell_{\mathrm{KL}}(f_{\btheta}(\bx), f_{\btheta}(\bx + \bdelta))\right ]$}}
    \label{eq: trades}
\tag{{\AdvRT}}
\end{align}
\vspace*{2mm}
}}%
where $\btheta$ denotes model parameters to be robustified, $(\mathbf x, y) \in \mathcal D$ is a training sample, drawn from the training set $\mathcal D$,  with input feature $\mathbf x$  and label $y$, $\ell(\btheta, \bx; y)$ denotes the cross-entropy loss using model $\btheta$  at data point $(\mathbf x, y) $, $\bdelta$ signifies the input perturbation variable subject to the $\ell_\infty$-norm ball of radius $\epsilon$,  
$f_{\btheta} (\cdot)$ denotes the model's predictions,
$\ell_{\mathrm{KL}}$ is the KL divergence loss that characterizes the worst-case prediction stability at the presence of $\boldsymbol \delta $, and $\lambda>0$ is a regularization parameter to strike the tradeoff between empirical risk minimization and the robustness of model predictions. 

Although {\AdvRT}  has been well studied for    adversarial robustness of CNNs, there exists few   attempts to robustify {\moe}.
This raises the problem of our interest:

\vspace*{1mm}
\begin{tcolorbox}[before skip=0.2cm, after skip=0.2cm, boxsep=0.0cm, middle=0.1cm, top=0.1cm, bottom=0.1cm]

\textbf{(Problem statement)} {Can   {\moe} be  robustified as effectively as an ordinary CNN   using \ref{eq: trades}? If not, how to robustly train   {\moe} to achieve   robustness not worse than  {\AdvRT}-oriented  {\dense}, {{\pcnn}}, and {\ldense}  while preserving MoE's efficiency? 
}
\end{tcolorbox}

\paragraph{Warm-up study: {\AdvRT} for {\moe} is \textit{not} trivial.}  Our goal to robustify {\moe} includes (1) achieving high robustness, (2) maintaining high prediction accuracy, and (3) making full use of  MoE routing to keep the model's high efficiency and expressiveness.
Nonetheless, the routing system in MoE brings extra robustification challenges, which never exist in ordinary CNNs. Specifically, the input-specific expert selection in MoE could make the attacker easier to succeed, since input perturbations can \textit{either} mislead routers to select   incorrect experts \textit{or} fool the pathway-designated predictor. Such a `\textit{two-way attack mode}' makes {\AdvRT} for {\moe} highly non-trivial.

\begin{wrapfigure}{r}{.5\columnwidth}
\vspace*{-2mm}
\centering
\hspace*{-3mm}\includegraphics[width=\linewidth]{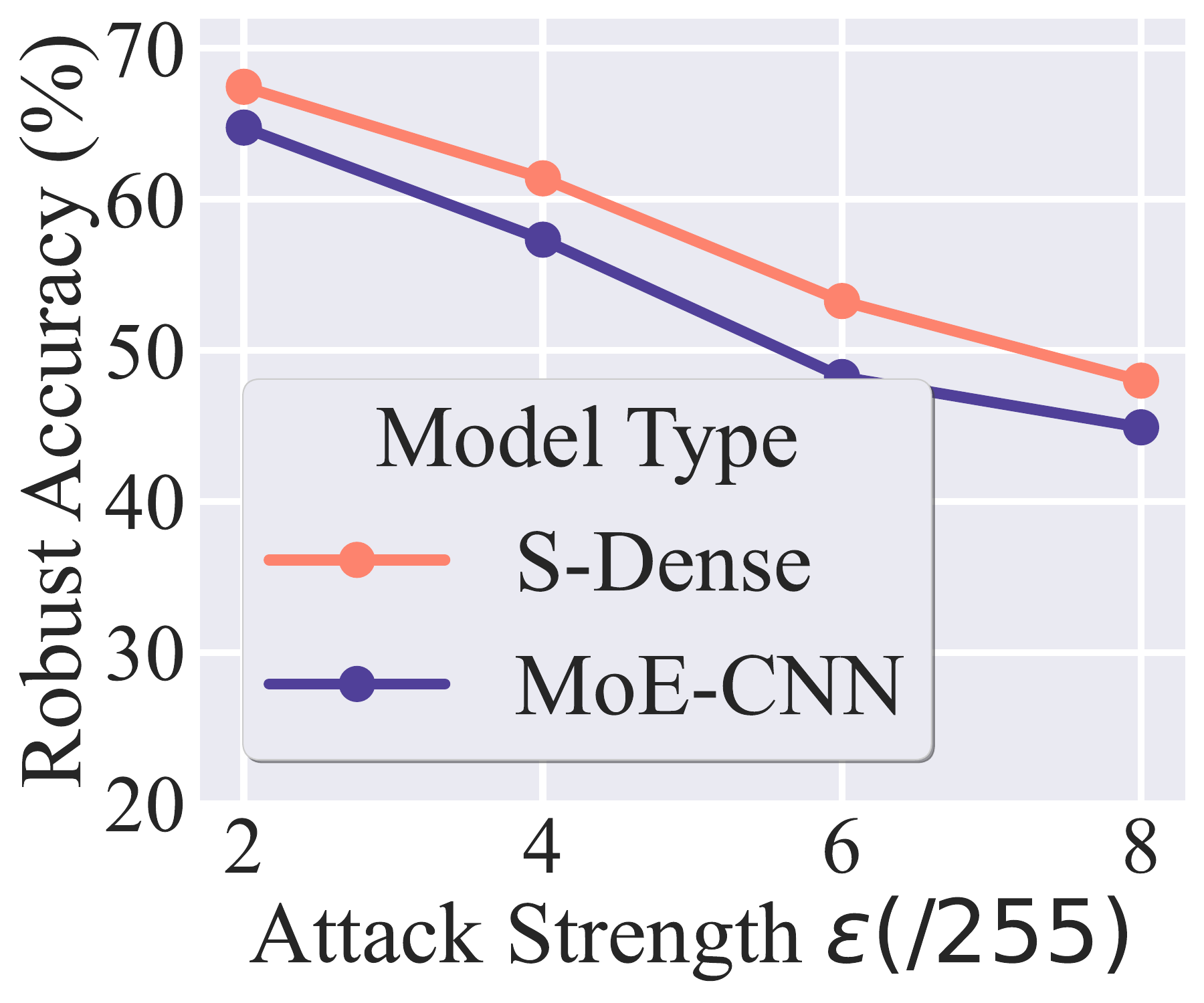}
\vspace*{-1mm}
 \caption{\footnotesize{Performance of {\moe} and {\dense} robustly trained   using \eqref{eq: trades} on {\cifarten} with ResNet-18 as the backbone.
}}
\label{fig: warm-up}
\vspace*{-3mm}
\end{wrapfigure}

\textbf{Fig.\,\ref{fig: warm-up}} empirically justifies that  the direct application of {\eqref{eq: trades}} to {\moe} is problematic. In Fig.\,\ref{fig: warm-up}, we consider ResNet-18 as the model backbone ({\ldense}) and {\cifarten} for image classification. We apply \eqref{eq: trades} to train {\moe} and {\dense}, and report the robust accuracy (RA), \textit{i.e.},  test-time  accuracy over adversarial examples generated by 50-step PGD attacks\,\cite{madry2018towards}, against different attack strengths $\epsilon$.
As we can see, although {\moe} has a much larger model capacity than {\dense}, it leads to a significant RA drop when the conventional {\AdvRT} approach is applied. This implies that the design of {\AdvRT} for {\moe} is far from trivial. A new robust learning protocol is thus needed to improve the robustness of {\moe} without losing its merits in efficiency and generalization.

\section{Methods}
\label{sec: methods}
In this section, we start by  {peering into} the failure case of \eqref{eq: trades} in {\moe}
by understanding the roles of the routers and  pathways in \eqref{eq: trades}. We empirically show that these individual components are hard to adapt to each other and cannot  make a concerted effort in {\AdvRT}. Based on that, we {develop} a new {\AdvRT} framework for {\moe},   {\ours}, which also takes inspiration from bi-level optimization.

\paragraph{Dissecting robustness of {\moe}: Routers' robustness vs. pathways' robustness.}
The main puzzle in robustifying {\moe} comes from the coupling between the robustness of routers (which are responsible for expert selection   across layers) and the robustness of the input-specific MoE pathways (which are in charge of the final prediction of an input).
Given the failure case of {\AdvRT} for {\moe} in {Fig.\,\ref{fig: warm-up}}, 
we need to understand the roles of routers and pathways in {\AdvRT}, \textit{i.e.}, how the adversarial robustness  of {\moe} is gained in the presence of the `two-way attack mode'.
To this end, we begin by assessing the influence of the routers' robustness on the overall robustness. This is also inspired by the recent pruning literature\,\cite{sehwag2020hydra} showing that model robustness can be gained solely from network's sparse topology (regardless of model weights). We thus ask:
\begin{center}
	\setlength\fboxrule{0.5pt}
	\noindent\fcolorbox{black}[rgb]{0.99,0.99,0.99}{\begin{minipage}{0.99\columnwidth}
        {\bf (Q1)} Is improving routers' robustness sufficient to achieve a robust {\moe}?
	\end{minipage}}
\end{center}

To tackle {\bf (Q1)}, we first split the parameters of {\moe} (\textit{i.e.}, $\btheta$) into two parts, the   parameters of routers $\boldsymbol \phi $ and the parameters of the backbone network $\boldsymbol \psi $. This yields $\btheta = [\boldsymbol \phi^\top, \boldsymbol \psi^\top]^\top$, where $\top$ is the transpose operation. We then call \eqref{eq: trades} to robustly train routers ($\boldsymbol \phi$) but \textit{fix} the backbone network ($\boldsymbol \psi$) at its standard pre-trained weights. 
We denote this partially-robustified model by $\bar{\btheta} = [\bar{\boldsymbol \phi}^\top, \boldsymbol \psi^\top]^\top$, where $\,\bar{}\,$ indicates the updated parameters.
To answer {\bf (Q1)}, we assess the   robustness gain of $\bar{\btheta}$ vs. 3 baselines (M1-M3): (\textbf{M1}) the standard {\moe} $\btheta$,
(\textbf{M2}) \ref{eq: trades}-robustified {\dense},  and (\textbf{M3}) {\pcnn} achieved by the robust sparse mask learning method  \cite{sehwag2020hydra} over the original {\ldense} model. 
\begin{wrapfigure}{r}{40mm}
\centering
\hspace*{-3mm}\includegraphics[width=.49\columnwidth]{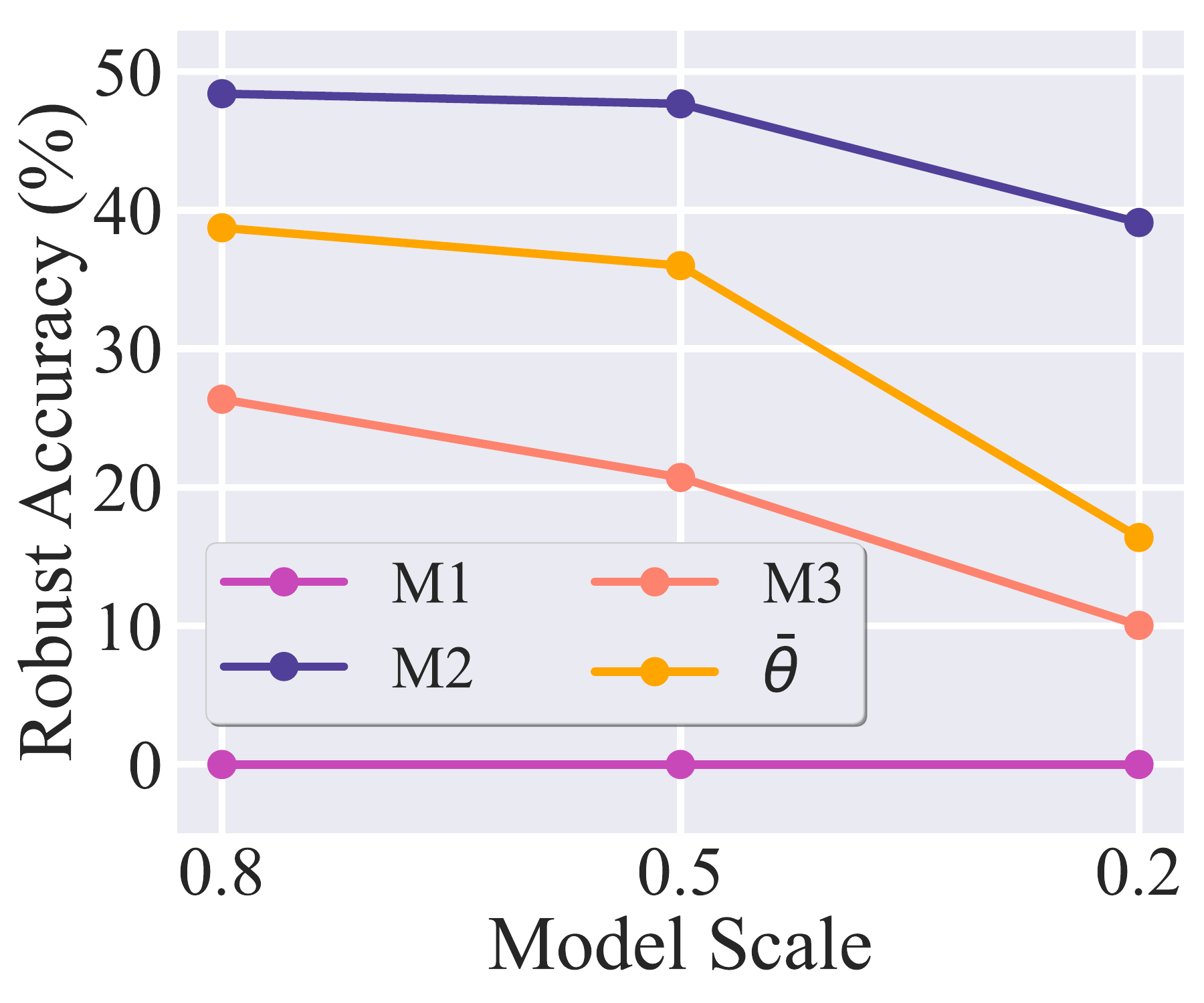}
\vspace*{-2mm}
\caption{{Robustness comparison of router-robusified {\moe} (\textit{i.e.}  $\bar{\btheta}$) and  baseline models (M1 -- M3) for different model scales under {\cifarten} given the backbone network ResNet-18.
    }}
    \label{fig: method_partI}
\end{wrapfigure}
\textbf{Fig.\,\ref{fig: method_partI}} shows the robust accuracy of  the router-robustified {\moe} $\bar{\btheta}$ and its performance comparison with other baseline models. 
As we can see, the robustified router improves the overall robustness (\textit{e.g.}, $37.64\%$ for $\bar{\btheta}$
with model scale $0.5$)
compared to the undefended {\moe} (\textbf{M1}: $0\%$) and the robustified mask (\textbf{M3}: $20.04\%$). However, there is still a huge robustness gap compared to the \eqref{eq: trades}-robustified {\dense} (\textbf{M2}: $47.68\%$). Based on the results above, we acquire the first insight into \textbf{(Q1)}:
\begin{center}
	\setlength\fboxrule{0.5pt}
	\noindent\fcolorbox{black}[rgb]{0.99,0.99,0.99}{\begin{minipage}{0.99\columnwidth}
        {\bf Insight\,1:}  Robustifying routers  improves the overall  robustness  of {\moe}   but is \textit{not} as effective as \ref{eq: trades}-resulted {\dense}.
	\end{minipage}}
\end{center}

Based on Insight\,1, we further peer into the resilience of expert selection decisions to adversarial examples. If expert selections in \textit{all} MoE layers keep intact in the presence of an adversarial perturbation, we say that the routing system of {\moe} is robust against this adversarial example. We then divide adversarial examples into \textbf{four categories} according to whether they successfully attacked routers and the router-oriented pathways: \ding{182}
\textit{unsuccessful} attack on \textit{both} routers and MoE pathways, \ding{183} \textit{successful} attack on routers but \textit{not MoE} pathways, \ding{184} \textit{successful} attack on MoE pathways but \textit{not routers}, and \ding{185} \textit{successful} attack on \textit{both} routers and MoE pathways. Here \ding{182} + \ding{184} characterizes the robustness of routers, while \ding{182} + \ding{183} represents that of MoE. Thus, if \ding{183} or \ding{184} takes a large portion of generated adversarial examples, it implies that the routers' robustness does \textit{not} directly impact the MoE pathway-based predictor's robustness. 
\begin{figure}[t]
    \centering    \includegraphics[width=.85\linewidth]{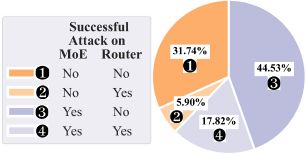}
    \caption{{Adversarial attack success analysis on dissected 
    {\moe} models ${\bar{\btheta}}=[\bar\bphi^\top, \bpsi^\top]$ (model scale $r=0.5$), where only $\bar\bphi$ is \eqref{eq: trades}-robustified. The adversarial evaluation is based on 50-step PGD attack\,\cite{madry2018towards} to fool ${\bar{\btheta}}$, and other experiment setups align with {Fig.\,\ref{fig: method_partI}}. The evaluation is carried out on the test set with a total number of $10000$ samples.
    }}
    \label{fig: router_analysis}
    \vspace*{-6mm}
\end{figure}
\textbf{Fig.\,\ref{fig: router_analysis} } shows the above categories \ding{182}-\ding{185} when attacking the router-robustified {\moe} (\textit{i.e.}, $\bar{\btheta}$). As we can see, routers' robustness indeed improves prediction robustness (as shown by $31.74\%$ unsuccessful attacks against the MoE predictor in \ding{182}). However, in the total number of unsuccessful attacks against routers (\textit{i.e.}, \ding{182}$+$\ding{184}$= 76.27\%$), more than half of them successfully fool the MoE predictor (\textit{i.e.}, \ding{184}$>$\ding{182}). The above results provide us an additional insight:
\begin{center}
\vspace*{-2.5mm}
	\setlength\fboxrule{0.5pt}
	\noindent\fcolorbox{black}[rgb]{0.99,0.99,0.99}{\begin{minipage}{0.99\columnwidth}
		{\bf Insight\,2:}  Improving routers' robustness is \textit{not} sufficient  for the MoE predictor  to gain satisfactory  robustness although the former makes a positive impact.
	\end{minipage}}
	\vspace*{-2.5mm}
\end{center}

Both \textbf{Insight\,1} and \textbf{Insight\,2} point out that only improving routers' robustness is \textit{not} adequate to obtain the desired robustness for the overall {\moe}.
Thus, we next ask:
\begin{center}
\vspace*{-2.5mm}
	\setlength\fboxrule{0.5pt}
	\noindent\fcolorbox{black}[rgb]{0.99,0.99,0.99}{\begin{minipage}{0.99\columnwidth}
	{\bf (Q2)} Given the router-robustified model $\bar{\btheta}$, can we equip $\bar{\btheta}$ with additional robustness by robustly training expert weights ($\boldsymbol \psi$)? And how does it further impact routers?
	\end{minipage}}
	\vspace*{-2.5mm}
\end{center}

\begin{wrapfigure}{r}{.5\columnwidth}
\centering
\hspace*{-.1mm}\includegraphics[width=.35\columnwidth]{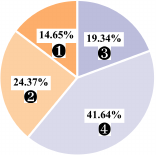}
\caption{{Adversarial attack success analysis on routers $\bar{\bphi}$ and {\moe} models ${\bar{\bar{\btheta}}}=[\bar\bphi^\top, \bar\bpsi^\top]$. Other setups remain the same as {Fig.\,\ref{fig: router_analysis}}.
}}
\label{fig: router_analysis_after}
\vspace*{-3mm}
\end{wrapfigure}
To answer {\bf (Q2)}, we call \eqref{eq: trades} to further robustly train the backbone network $\boldsymbol \psi$ on top of the router-robustified model $\bar{\btheta}$. We denote the resulting model by $\bar{\bar{\btheta}} = [\bar{\boldsymbol \phi}^\top, \bar{\boldsymbol \psi}^\top]$. 
\textbf{Fig.\,\ref{fig: router_analysis_after}} shows the dissection of the robustness of $\bar{\bar{\btheta}}$ in the same setup of {Fig.\,\ref{fig: router_analysis}}. 
Obviously, the overall prediction robustness {(\ding{182}+\ding{183})} is further enhanced after updating ${\bar{\btheta}}$ to $\bar{\bar{\btheta}}$. Thus, the gains in the robustness of experts' weights indeed further help improve the overall robustness.
However, this   leads to a surprising drop in the router's robustness ({\ding{182}+\ding{184}}) when comparing $\bar{\bar{\btheta}}$ with ${\bar{\btheta}}$. 
This shows that routers' robustness is \textit{not} automatically preserved if experts are updated. We obtain the following insight into \textbf{(Q2)}:
\begin{center}
\vspace*{-2mm}
	\setlength\fboxrule{0.5pt}
	\noindent\fcolorbox{black}[rgb]{0.99,0.99,0.99}{\begin{minipage}{0.99\columnwidth}
				{\bf Insight\,3:} Robustifying routers and MoE weights can yield complementary   benefits but  the inadaptability of routers' robustness to MoE's robustness prevents {\AdvRT} from achieving significant robustness improvement.
	\end{minipage}}
\end{center}

\noindent\textbf{{\ours}: Router-expert alternating {\AdvRT} through a bi-level optimization viewpoint.}
As illuminated by insights above, we provide a reason for the ineffectiveness of \ref{eq: trades} in robustifying {\moe}.
\textbf{Insights}\,\textbf{1-2} show that  the robustness of routers $(\boldsymbol \phi)$ and the robustness of MoE-based predictor $(\boldsymbol \psi)$  are intertwined and their interrelation is non-trivial. As a result,  the single-level (non-convex)  robust optimization over  the entire model parameters $(\boldsymbol \phi, \boldsymbol \psi)$  experiences difficulty in co-optimizing routers and MoE prediction pathways to achieve the best complementary robustness gains, as supported by \textbf{Insight\,3}. A key missing optimization factor in {\AdvRT} for {\moe} is its incapability of modeling and optimizing the coupling between the robustness of the routers and that of the MoE pathways. Without such an   optimization design, it is difficult for \ref{eq: trades} to robustify routers and MoE  pathways in a cooperative and adaptive mode.

Spurred by the above, we develop a new {\AdvRT} framework through bi-level optimization (\textbf{BLO}). In general, BLO provides a   hierarchical learning framework with two levels of optimization tasks, where the objective and variables of an upper-level problem depend on the optimizer of the lower level. BLO then enables us to explicitly model the coupling between {\AdvRT} for routers and {\AdvRT} for MoE network. Specifically, we modify the conventional \eqref{eq: trades} to 

\vspace*{-4mm}
{\small{ \begin{align}
\begin{array}{ll}
  \displaystyle \minimize_{\boldsymbol \psi}       & \ell_\mathrm{TRADES} (\boldsymbol \psi, \boldsymbol \phi^*(\boldsymbol \psi)  ;\mathcal{D})\\
   \st       & \boldsymbol \phi^*(\boldsymbol \psi)  = \argmin_{\boldsymbol \phi} \ell_\mathrm{TRADES} (\boldsymbol \psi, \boldsymbol \phi ;\mathcal{D}),
    \end{array}
    \label{eq: advMoE_prob}
\end{align}}}%
where the   model  parameters of {\moe} $\boldsymbol \theta$  are split into the {lower-level} optimization variables $\boldsymbol \phi$ for routers and the {upper-level} optimization variables $\boldsymbol \psi$  for the MoE backbone network, and  $\ell_\mathrm{TRADES} (\boldsymbol \psi, \boldsymbol \phi ;\mathcal{D})$ denotes the {\trades}-type training loss defined in \eqref{eq: trades} by replacing $\btheta$ with $(\boldsymbol \phi, \boldsymbol \psi )$. 
Compared to \eqref{eq: trades}, our proposal \eqref{eq: advMoE_prob} has the following differences. \textbf{First}, robustifying MoE network ($\boldsymbol \psi$) is now explicitly coupled with   routers' optimization   through the lower-level solution $\boldsymbol \phi^*(\boldsymbol \psi)$. \textbf{Second}, our proposal addresses the robustness adaptation problem  pointed out in \textbf{Insight\,3} since the lower-level optimization of \eqref{eq: advMoE_prob} enables fast   adaptation of $\boldsymbol \phi$ to the current MoE network $\boldsymbol \psi$ like meta-learning \cite{finn2017model}. \textbf{Third}, since $\ell_\mathrm{TRADES}$ is involved at both optimization levels of \eqref{eq: advMoE_prob}, the embedded attack generation problem (\textit{i.e.}, maximization over $\boldsymbol \delta$ in \ref{eq: trades}) needs to be solved at each level but   corresponding to different victim models, \textit{i.e.}, $(\boldsymbol \psi, \boldsymbol \phi)$ and $(\boldsymbol \psi, \boldsymbol \phi^*(\boldsymbol \psi))$, respectively.

\begin{algorithm}[htb]
\caption{The {\ours} algorithm}
\begin{algorithmic}[1]
\small{
\State \textbf{Initialize:} backbone network $\bpsi$, routers $\bphi$, batch size $b$, attack generation step $K$.
\For{Iteration $t=0,1,\ldots,$}
 \State Pick {different} random data batches $\mathcal B_{\psi}$ and $\mathcal B_{\phi}$ for 
 \hspace*{5.5mm}backbone and router training
\State Lower-level $\bphi$-update (with fixed $\bpsi$):
Given $\boldsymbol \psi$, update $\bphi$ 
\hspace*{4.8mm}by minimizing $\ell_{\mathrm{TRADES}}$  using    $K$-step PGD attack\,\cite{madry2018towards} \hspace*{4mm} generator  and SGD (with   $\mathcal{B}_{\bphi}$)
\State Upper-level $\bpsi$-update (with fixed $\bphi$):
Given   $\bphi$, update $\bpsi$ 
\hspace*{4mm} by minimizing $\ell_{\mathrm{TRADES}}$  using   $K$-step PGD attack gen-
\hspace*{4mm}-erator  and SGD (with   $\mathcal{B}_{\bpsi}$) 
\EndFor}
\end{algorithmic}
\label{alg: advmoe}
\end{algorithm}

To solve problem \eqref{eq: advMoE_prob}, we adopt a standard alternating optimization (AO) method \cite{bezdek2003convergence}.  Compared with other kinds of BLO algorithms \cite{liu2021investigating},  AO is the most computationally efficient. Our extensive experiments in Sec.\,\ref{sec: exp} will show that AO is  effective to  boost the adversarial robustness of {\moe} and achieve improvements over baseline methods and models by a substantial margin.
The key idea of AO is to alternatively optimize the lower-level and the upper-level problem, during which variables defined in another level are fixed. We term the resulting algorithmic framework as \underline{Adv}ersarially robust learning for \underline{MoE}-CNN (\ours); see Algorithm\,\ref{alg: advmoe} for a summary.

We highlight that {\ours} will train robust routers and robust MoE pathways to `accommodate' each other. In contrast to the conventional \ref{eq: trades} framework, {\ours} delivers the coupled $\boldsymbol \phi^*(\boldsymbol \psi)$ and $\boldsymbol \psi$, where both parts make a concerted effort to improve the overall robustness. We also remark that {\ours} does not introduce additional hyper-parameters, since in practice we found routers and experts can share the same learning rate and schedules. More implementation details are provided in Appendix\,\ref{app:training}. {In the meantime, we remark that since our proposal is a BLO with non-convex lower and upper-level objectives \eqref{eq: advMoE_prob}. It is difficult to prove the convergence of {\ours}. Existing theoretical analysis of BLO typically relies    on  strongly convex
assumptions of lower-level problems \cite{hong2020two, zhang2023introduction}. Although without a proper theoretical analysis framework,  our method converges well in practice (see Appendix\,\ref{app:more_exps}).}

\begin{table*}[htb]
\centering
\definecolor{rulecolor}{RGB}{0,71,171}
\definecolor{tableheadcolor}{RGB}{204,229,255}
\newcommand{\myrowcolour}{\rowcolor{tableheadcolor}}
\newcommand{\highest}[1]{\textcolor{blue}{\textbf{#1}}}
\newcommand{\topline}{ %
    \arrayrulecolor{rulecolor}\specialrule{0.1em}{\abovetopsep}{0pt}%
    \arrayrulecolor{rulecolor}\specialrule{\lightrulewidth}{0pt}{0pt}%
    \arrayrulecolor{tableheadcolor}\specialrule{\aboverulesep}{0pt}{0pt}%
    \arrayrulecolor{rulecolor}
    }
\newcommand{\midtopline}{
    \arrayrulecolor{tableheadcolor}\specialrule{\aboverulesep}{0pt}{0pt}%
    \arrayrulecolor{rulecolor}\specialrule{\lightrulewidth}{0pt}{0pt}%
    \arrayrulecolor{white}\specialrule{\aboverulesep}{0pt}{0pt}%
    \arrayrulecolor{rulecolor}}
    \newcommand{\bottomline}{
    \arrayrulecolor{tableheadcolor}\specialrule{\aboverulesep}{0pt}{0pt}
    \arrayrulecolor{rulecolor}
    \specialrule{\heavyrulewidth}{0pt}{\belowbottomsep}
    \arrayrulecolor{rulecolor}\specialrule{\lightrulewidth}{0pt}{0pt}
    }
\newcolumntype{?}{!{\vrule width 1.4pt}}
\caption{{{Performance overview of {\ours} (our proposal) \textit{vs.} baselines on various datasets and model backbone architectures. The model scale is fixed at $r=0.5$ for Dense-CNN, Sparse-CNN and Moe-CNN (denoted with the symbol {\partcirc}, since they are fairly comparable to each other) compared to {\ldense} ($r=1.0$, denoted with the symbol {\fullcirc}). For train- and test-time attack generations, we adopt an attack strength of $\epsilon=8/255$ for {\cifarten} and {\cifarhun}, and $\epsilon=2/255$ for {\timagenet} and {\imagenet}. We evaluate RA (robust test accuracy) against 50-step PGD attack\,\cite{madry2018towards}, SA (standard test accuracy), and GFLOPS (FLOPS $\times 10^9$) per test-time example (test-time inference efficiency) for each model. In each (dataset, backbone) setup, \ding{172} we highlight the best SA and RA over all baselines per model scale in \textbf{bold}, and \ding{173} we mark the performance better than {\denseRT} in \colorbox{green!30}{green}. Results in the format of $a${\footnotesize{$\pm b$}} provide the mean value $a$ and its standard deviation $b$ over $3$ independent trials.}}}
\label{tab: exp_main}
\resizebox{\textwidth}{!}{%
\begin{tabular}{l|c|ccc?l|c|ccc}
\topline
\textbf{Method} &
\textbf{Backbone} &
  \textbf{RA} (\%) &
  \textbf{SA} (\%) &
  \textbf{GFLOPS}(\#) &
  \textbf{Method} &
  \textbf{Backbone} &
  \textbf{RA} (\%) &
  \textbf{SA} (\%) &
  \textbf{GFLOPS} (\#)
\\ 
\midtopline \myrowcolour
\multicolumn{10}{c}{\textbf{\texttt{CIFAR-10}}} \\ \midtopline
  {\fullcirc}{\denseRT}  
  & \multirow{5}{*}{ResNet-18} 
  & 50.13\footnotesize{$\pm 0.13$}
  & 82.99\footnotesize{$\pm 0.11$}
  & 0.54
  
  & {\fullcirc}{\denseRT}  
  & \multirow{5}{*}{WRN-28-10} 
  & 51.75\footnotesize{$\pm 0.12$}
  & 83.54\footnotesize{$\pm 0.15$}
  & 5.25 
\\ 
  {\partcirc}{\slimRT} 
  & 
  & 48.12\footnotesize{$\pm 0.09$}
  & 80.18\footnotesize{$\pm 0.11$}
  & 0.14 (74\%$\downarrow$)
  
  & {\partcirc}{\slimRT} 
  & 
  & 50.66\footnotesize{$\pm 0.13$}
  & 82.24\footnotesize{$\pm 0.10$}
  & 1.31 (75\%$\downarrow$)
  
\\
  {\partcirc}{\sparseRT} 
  &
  & 47.93\footnotesize{$\pm 0.17$}
  & \textbf{80.45}\footnotesize{$\pm 0.13$}
  & 0.14 (74\%$\downarrow$)
  
  & {\partcirc}{\sparseRT} 
  & 
  & 48.95\footnotesize{$\pm 0.14$}
  & 82.44\footnotesize{$\pm 0.17$}
  & 1.31 (75\%$\downarrow$)
\\
  {\partcirc}{\moeRT} 
  & 
  & 45.57\footnotesize{$\pm 0.51$}
  & 78.84\footnotesize{$\pm 0.75$}
  & 0.15 (72\%$\downarrow$)
  & {\partcirc}{\moeRT} 
  &
  & 46.73\footnotesize{$\pm 0.46$}
  & 77.42\footnotesize{$\pm 0.73$}
  & 1.75 (67\%$\downarrow$)
\\
\gr {\partcirc}\textbf{{\ours}} 
  &
  & \colorbox{green!30}{\textbf{51.83}}\footnotesize{$\pm 0.12$}
  & 80.15\footnotesize{$\pm 0.11$}
  & 0.15 (72\%$\downarrow$)
  
  & {\partcirc}\textbf{{\ours}} 
  &
  & \colorbox{green!30}{\textbf{55.73}}\footnotesize{$\pm 0.13$}
  & \colorbox{green!30}{\textbf{84.32}}\footnotesize{$\pm 0.18$}
  & 1.75 (67\%$\downarrow$)
  
\\ \midtopline
  {\fullcirc}{\denseRT}  
  &\multirow{5}{*}{VGG-16} 
  & 46.19\footnotesize{$\pm 0.21$}
  & 82.18\footnotesize{$\pm 0.23$}
  & 0.31
  
  & {\fullcirc}{\denseRT}  
  & \multirow{5}{*}{DenseNet} 
  & 44.52\footnotesize{$\pm 0.14$}
  & 74.97\footnotesize{$\pm 0.19$}
  & 0.07
  
\\
{\partcirc}{\slimRT} 
  &
  & 45.72\footnotesize{$\pm 0.18$}
  & \textbf{80.10}\footnotesize{$\pm 0.16$}
  & 0.07 (77\%$\downarrow$)
  
  & {\partcirc}{\slimRT} 
  &
  & 38.07\footnotesize{$\pm 0.13$}
  & 69.63\footnotesize{$\pm 0.11$}
  & 0.02 (71\%$\downarrow$)
  
\\
  {\partcirc}{\sparseRT} 
  &
  & 46.13\footnotesize{$\pm 0.15$}
  & 79.32\footnotesize{$\pm 0.18$}
  & 0.07 (77\%$\downarrow$)
  & {\partcirc}{\sparseRT} 
  &
  & 37.73\footnotesize{$\pm 0.13$}
  & 67.35\footnotesize{$\pm 0.12$}
  & 0.02 (71\%$\downarrow$)
\\
  {\partcirc}{\moeRT} 
  &
  & 43.37\footnotesize{$\pm 0.46$}
  & 76.49\footnotesize{$\pm 0.65$}
  & 0.12 (61\%$\downarrow$)
  
  & {\partcirc}{\moeRT} 
  &
  & 35.21\footnotesize{$\pm 0.74$}
  & 64.41\footnotesize{$\pm 0.81$}
  & 0.03 (57\%$\downarrow$)
  
\\
\gr{\partcirc}\textbf{{\ours}} 
  &
  & \colorbox{green!30}{\textbf{49.82}}\footnotesize{$\pm 0.11$}
  & 80.03\footnotesize{$\pm 0.10$}
  & 0.12 (61\%$\downarrow$)
  
  & {\partcirc}\textbf{{\ours}} 
  & 
  & \textbf{39.97}\footnotesize{$\pm 0.11$}
  & \textbf{70.13}\footnotesize{$\pm 0.15$}
  & 0.03 (57\%$\downarrow$)
\\
\midtopline \myrowcolour
\multicolumn{10}{c}{\textbf{\texttt{CIFAR-100}}} \\ \midtopline
  {\fullcirc}{\denseRT}  
  & \multirow{5}{*}{ResNet-18} 
  & 27.23\footnotesize{$\pm 0.08$}
  & 58.21\footnotesize{$\pm 0.12$}
  & 0.54
  
  & {\fullcirc}{\denseRT}  
  & \multirow{5}{*}{WRN-28-10} 
  & 27.90\footnotesize{$\pm 0.13$}
  & 57.60\footnotesize{$\pm 0.09$}
  & 5.25 
\\
  {\partcirc}{\slimRT} 
  &
  & 26.41\footnotesize{$\pm 0.16$}
  & 57.02\footnotesize{$\pm 0.14$}
  & 0.14 (74\%$\downarrow$)
  
  & {\partcirc}{\slimRT} 
  &
  & 26.30\footnotesize{$\pm 0.10$}
  & 56.80\footnotesize{$\pm 0.08$}
  & 1.31 (75\%$\downarrow$)
  
\\
  {\partcirc}{\sparseRT} 
  &
  & 26.13\footnotesize{$\pm 0.14$}
  & 57.24\footnotesize{$\pm 0.12$}
  & 0.14 (74\%$\downarrow$)
  
  & {\partcirc}{\sparseRT} 
  & 
  & 25.83\footnotesize{$\pm 0.16$}
  & 57.39\footnotesize{$\pm 0.14$}
  & 1.31 (75\%$\downarrow$)
\\
  {\partcirc}{\moeRT} 
  &
  & 22.72\footnotesize{$\pm 0.42$}
  & 53.34\footnotesize{$\pm 0.61$}
  & 0.15 (72\%$\downarrow$)
  & {\partcirc}{\moeRT} 
  & 
  & 22.94\footnotesize{$\pm 0.55$}
  & 53.39\footnotesize{$\pm 0.49$}
  & 1.75 (67\%$\downarrow$)
\\
\gr {\partcirc}\textbf{{\ours}} 
  &
  & \colorbox{green!30}{\textbf{28.05}}\footnotesize{$\pm 0.13$}
  & \textbf{57.73}\footnotesize{$\pm 0.11$}
  & 0.15 (72\%$\downarrow$)
  
  & {\partcirc}\textbf{{\ours}} 
  &
  & \colorbox{green!30}{\textbf{28.82}}\footnotesize{$\pm 0.14$}
  & \textbf{57.56}\footnotesize{$\pm 0.17$}
  & 1.75 (67\%$\downarrow$)
  
\\ \midtopline
  {\fullcirc}{\denseRT}  
  &\multirow{5}{*}{VGG-16} 
  & 22.37\footnotesize{$\pm 0.15$}
  & 52.36\footnotesize{$\pm 0.17$}
  & 0.31
  
  & {\fullcirc}{\denseRT}  
  & \multirow{5}{*}{DenseNet} 
  & 21.72\footnotesize{$\pm 0.13$}
  & 48.64\footnotesize{$\pm 0.14$}
  & 0.07
  
\\
{\partcirc}{\slimRT} 
  & 
  & 20.58\footnotesize{$\pm 0.13$}
  & \textbf{48.89}\footnotesize{$\pm 0.14$}
  & 0.07 (77\%$\downarrow$)
  
  & {\partcirc}{\slimRT} 
  &
  & 16.86\footnotesize{$\pm 0.21$}
  & 39.97\footnotesize{$\pm 0.11$}
  & 0.02 (71\%$\downarrow$)
  
\\
  {\partcirc}{\sparseRT} 
  &
  & 21.12\footnotesize{$\pm 0.22$}
  & 48.03\footnotesize{$\pm 0.17$}
  & 0.07 (77\%$\downarrow$)
  & {\partcirc}{\sparseRT} 
  &
  & 17.72\footnotesize{$\pm 0.14$}
  & 41.03\footnotesize{$\pm 0.16$}
  & 0.02 (71\%$\downarrow$)
\\
  {\partcirc}{\moeRT} 
  &
  & 19.34\footnotesize{$\pm 0.43$}
  & 45.51\footnotesize{$\pm 0.75$}
  & 0.12 (61\%$\downarrow$)
  
  & {\partcirc}{\moeRT} 
  &
  & 14.45\footnotesize{$\pm 0.45$}
  & 36.72\footnotesize{$\pm 0.71$}
  & 0.03 (57\%$\downarrow$)
  
\\
\gr{\partcirc}\textbf{{\ours}} 
  & 
  & \textbf{21.21}\footnotesize{$\pm 0.21$}
  & 48.33\footnotesize{$\pm 0.17$}
  & 0.12 (61\%$\downarrow$)
  
  & {\partcirc}\textbf{{\ours}} 
  &
  & \colorbox{green!30}{\textbf{23.31}}\footnotesize{$\pm 0.11$}
  & \colorbox{green!30}{\textbf{48.97}}\footnotesize{$\pm 0.14$}
  & 0.03 (57\%$\downarrow$)
\\
\midtopline \myrowcolour
\multicolumn{10}{c}{\textbf{\texttt{Tiny-ImageNet}}} \\ \midtopline
  
  {\fullcirc}{\denseRT}  
  & \multirow{5}{*}{ResNet-18} 
  & 38.17\footnotesize{$\pm 0.14$}
  & 53.81\footnotesize{$\pm 0.16$}
  & 2.23
  
  & {\fullcirc}{\denseRT}  
  & \multirow{5}{*}{WRN-28-10} 
  & 38.82\footnotesize{$\pm 0.15$}
  & 55.30\footnotesize{$\pm 0.19$}
  & 21.0
\\
  {\partcirc}{\slimRT} 
  &
  & 36.29\footnotesize{$\pm 0.16$}
  & 52.15\footnotesize{$\pm 0.13$}
  & 0.55 (75\%$\downarrow$)
  
  & {\partcirc}{\slimRT} 
  & 
  & 37.09\footnotesize{$\pm 0.12$}
  & 54.83\footnotesize{$\pm 0.16$}
  & 5.26 (75\%$\downarrow$)
  
\\
  {\partcirc}{\sparseRT} 
  &
  & 36.11\footnotesize{$\pm 0.13$}
  & 50.75\footnotesize{$\pm 0.17$}
  & 0.55 (75\%$\downarrow$)
  
  & {\partcirc}{\sparseRT} 
  &
  & 37.32\footnotesize{$\pm 0.14$}
  & 54.32\footnotesize{$\pm 0.23$}
  & 5.26 (75\%$\downarrow$)
\\

  {\partcirc}{\moeRT} 
  &
  & 34.41\footnotesize{$\pm 0.31$}
  & 47.73\footnotesize{$\pm 0.41$}
  & 0.75 (68\%$\downarrow$)
  
  & {\partcirc}{\moeRT} 
  &
  & 33.31\footnotesize{$\pm 0.41$}
  & 49.91\footnotesize{$\pm 0.52$}
  & 7.44 (65\%$\downarrow$)
  
\\
\gr {\partcirc}\textbf{{\ours}} 
  &
  & \colorbox{green!30}{\textbf{39.99}}\footnotesize{$\pm 0.12$}
  & \textbf{53.31}\footnotesize{$\pm 0.14$}
  & 0.75 (68\%$\downarrow$)
  
  & {\partcirc}\textbf{{\ours}} 
  &
  & \colorbox{green!30}{\textbf{40.15}}\footnotesize{$\pm 0.15$}
  & \textbf{55.18}\footnotesize{$\pm 0.09$}
  & 7.44 (65\%$\downarrow$)
  
\\ 
\midtopline \myrowcolour
\multicolumn{10}{c}{\textbf{\texttt{ImageNet}}} \\ \midtopline
  {\fullcirc}{\denseRT}  
  & \multirow{5}{*}{ResNet-18} 
  & 44.64\footnotesize{$\pm 0.14$}
  & 60.32\footnotesize{$\pm 0.15$}
  & 1.82
  
  & {\fullcirc}{\denseRT}  
  & \multirow{5}{*}{WRN-28-10} 
  & 45.13\footnotesize{$\pm 0.14$}
  & 60.97\footnotesize{$\pm 0.16$}
  & 16.1
  
\\
  {\partcirc}{\slimRT} 
  &
  & 41.19\footnotesize{$\pm 0.16$}
  & 58.32\footnotesize{$\pm 0.12$}
  & 0.48 (74\%$\downarrow$)
  & {\partcirc}{\slimRT} 
  & 
  & 41.72\footnotesize{$\pm 0.15$}
  & 58.98\footnotesize{$\pm 0.18$}
  & 4.04 (75\%$\downarrow$)
  
\\
  {\partcirc}{\sparseRT} 
  &
  & 40.87\footnotesize{$\pm 0.15$}
  & 58.22\footnotesize{$\pm 0.13$}
  & 0.48 (74\%$\downarrow$)
  
  & {\partcirc}{\sparseRT} 
  & 
  & 39.88\footnotesize{$\pm 0.18$}
  & \textbf{59.21}\footnotesize{$\pm 0.14$}
  & 4.04 (75\%$\downarrow$)
  
\\
  {\partcirc}{\moeRT} 
  &
  & 35.57\footnotesize{$\pm 0.73$}
  & 55.47\footnotesize{$\pm 0.66$}
  & 0.67 (63\%$\downarrow$)
  
  & {\partcirc}{\moeRT} 
  & 
  & 37.42\footnotesize{$\pm 0.44$}
  & 56.44\footnotesize{$\pm 0.71$}
  & 5.15 (68\%$\downarrow$)
  
\\
\gr
  {\partcirc}\textbf{{\ours}} 
  &
  & \colorbox{green!30}{\textbf{43.32}}\footnotesize{$\pm 0.12$}
  & \textbf{59.72}\footnotesize{$\pm 0.17$}
  & 0.67 (63\%$\downarrow$)
  
  & {\partcirc}\textbf{{\ours}} 
  &
  & \colorbox{green!30}{\textbf{46.82}}\footnotesize{$\pm 0.11$}
  & 58.87\footnotesize{$\pm 0.07$}
  & 5.15 (68\%$\downarrow$)
  
\\ 
\midtopline
\bottomline
\end{tabular}%
}
\vspace*{-5mm}
\end{table*}

\section{Experiments}
\label{sec: exp}
In this section, we will demonstrate the effectiveness of our proposed {\ours} approach on diverse datasets and models. 
We will also make an in-depth analysis of the router utility and the expert selection distribution for {\ours}-trained {\moe}. 

\subsection{Experiment Setup}
\label{sec: exp_setup}

\paragraph{Model and dataset setups.} 
To implement {\moe} and other baselines, we conduct experiments on ResNet-18\,\cite{he2016deep}, Wide-ResNet-28-10\,\cite{zagoruyko2016wide}, VGG-16\,\cite{simonyan2014very}, and DenseNet\,\cite{huang2017densely}. Towards fair assessment, our performance comparison between different model types is restricted to using {the same model scale} parameter $r$ (see Fig.\,\ref{fig: overview} for an example). By doing so,   an input example will leverage the same amount of model parameters for decision-making. For {\moe}, we consider $N=2$ experts with $r=0.5$ by default, see Appendix~\ref{app:training} for more details.
Dataset-wise, we focus on the commonly used ones to evaluate the adversarial robustness of image classification\,\cite{madry2018towards, zhang2019theoretically, zhang2021revisiting}, including {\cifarten}\,\cite{Krizhevsky2009learning}, {\cifarhun}\,\cite{Krizhevsky2009learning}, {\timagenet}\,\cite{deng2009imagenet}, and {\imagenet}\,\cite{deng2009imagenet}. 

\paragraph{Baselines.}
To make our performance comparison informative and comprehensive, we   consider three kinds of baselines that are fairly comparable to ({\ours}). 
\ding{172} {\slimRT}: we apply \ref{eq: trades} to {\dense}; \ding{173} {\sparseRT}:  we apply the robustness-aware (structured) sparse mask learning method \cite{sehwag2020hydra} to obtain {\pcnn}; \ding{174} {\moeRT}: we directly apply \ref{eq: trades} to {\moe}, which co-trains the routers and backbone network. Note this method is also adopted in the latest robust training  algorithm\,\cite{puigcerver2022adversarial} for ViT-based MoE architectures. 
It is worth noting that the above baselines use the same number of model parameters as the pathway of {\moe} during model prediction. 
In addition, we cover \ding{175} {\denseRT} (applying \ref{eq: trades} to {\ldense}) to acquire a robustness performance reference. Yet, we remark that it is \textit{not} quite fair to directly compare {\ldense} with {the aforementioned smaller counterparts}, since the former uses a larger model scale ($r=1.0$) at test-time inference.

\noindent\paragraph{Training and evaluation.}
We use TRADES\,\cite{zhang2019theoretically} as the default robust training objective for all baselines. We also follow the   literature\,\cite{madry2018towards, wong2020fast, zhang2019theoretically, zhang2021revisiting} to set the attack strength by $\epsilon = 8/255$ for {\cifarten} and {\cifarhun}, and $\epsilon=2/255$ for {\timagenet} and {\imagenet}.
To implement {\ours} (Algorithm\,1), we mimic the TRADES training pipeline but conduct the proposed BLO routine   to robustify routers and backbone parameters in an interactive mode. We adopt $2$-step PGD attack\,\cite{madry2018towards} at training time for \textit{all} the methods, supported by the recent work\,\cite{zhang2022revisiting} showing its compelling performance in {\AdvRT}.
We refer readers to Appendix~\ref{app:training} for more training details. During  evaluation, we report  standard accuracy (\textbf{SA}) on the clean test dataset and robust accuracy (\textbf{RA}) against test-time 
  50-step PGD attacks\,\cite{madry2018towards} with the attack strength same as the training values.
We also report \textbf{GFLOPs} (FLOPS $\times 10^9$) as an indicator of the test-time inference efficiency.

\subsection{Experiment Results}
\label{sec: exp_results}

\paragraph{Overall performance.} 
\textbf{Tab.\,\ref{tab: exp_main}} presents the overall performance of our proposed {\ours} algorithm vs. baselines. We make several key observations below. 

\underline{\textbf{First}}, {\ours} yields a significant robustness enhancement over all the baselines in every data-model setup.
Specifically, {\ours} consistently yields an improvement of around $1\% \sim 5\%$  on the robustness measured by RA against PGD attacks. 
Notably, {\ours} can also outperform  {\fullcirc}{\denseRT} in most cases,   around $1\% \sim 4\%$ robustness improvement (see highlighted   results in \colorbox{green!30}{green}).
This is remarkable since {\ldense} ($r=1.0$) is twice larger than an MoE pathway ($r=0.5$).
\underline{\textbf{Second}}, we observe that {\ours} has a preference on wider models. For instance, when  WRN-28-10  (the widest model architecture in experiments) is used,
{\ours} yields better robustness over the {\ldense} counterpart across all the dataset setups.
\underline{\textbf{Third}}, we also observe that the direct \ref{eq: trades} application to {\moe}, \textit{i.e.}, {\moeRT}, is worse than {\slimRT} and   {\ours} in all setups. This is consistent with our findings in Sec.\,\ref{sec: methods}. We remark that although the usefulness of {\moeRT} was exploited in \cite{puigcerver2022adversarial} for the MoE-type ViT, it is \textit{not} effective for training MoE-type CNNs anymore.
\underline{\textbf{Fourth}}, {\ours} can retain the high inference efficiency for {\moe}, as evidenced by the GFLOPS measurements in Tab.\,\ref{tab: exp_main}.  Compared to {\dense}, {\moe}  introduces minor computational overhead due to the routing system. However, it saves more than $50\%$ of the inference cost  vs. {\ldense}. 
This implies that {our proposal} {\ours} can preserve the efficiency merit of the {MoE} structure while effectively improving its adversarial robustness.

\begin{table*}[t]
\centering
\definecolor{rulecolor}{RGB}{0,71,171}
\definecolor{tableheadcolor}{RGB}{204,229,255}
\newcommand{\myrowcolour}{\rowcolor{tableheadcolor}}
\newcommand{\highest}[1]{\textcolor{blue}{\textbf{#1}}}
\newcommand{\topline}{ %
    \arrayrulecolor{rulecolor}\specialrule{0.1em}{\abovetopsep}{0pt}%
    \arrayrulecolor{rulecolor}\specialrule{\lightrulewidth}{0pt}{0pt}%
    \arrayrulecolor{tableheadcolor}\specialrule{\aboverulesep}{0pt}{0pt}%
    \arrayrulecolor{rulecolor}
    }
\newcommand{\midtopline}{
    \arrayrulecolor{tableheadcolor}\specialrule{\aboverulesep}{0pt}{0pt}%
    \arrayrulecolor{rulecolor}\specialrule{\lightrulewidth}{0pt}{0pt}%
    \arrayrulecolor{white}\specialrule{\aboverulesep}{0pt}{0pt}%
    \arrayrulecolor{rulecolor}}
    \newcommand{\bottomline}{
    \arrayrulecolor{tableheadcolor}\specialrule{\aboverulesep}{0pt}{0pt}
    \arrayrulecolor{rulecolor}
    \specialrule{\heavyrulewidth}{0pt}{\belowbottomsep}
    \arrayrulecolor{rulecolor}\specialrule{\lightrulewidth}{0pt}{0pt}
    }
\newcolumntype{?}{!{\vrule width 1.4pt}}
\caption{{{Robustness overview evaluated with AutoAttack\,\cite{croce2020reliable} (\textbf{RA-AA}) on various datasets and model backbone architectures. Other settings strictly follow Tab.\,\ref{tab: exp_main}. The values of RA-PGD, SA, and GFLOPS are repeated from Tab.\,\ref{tab: exp_main} for better comparison.
}}}
\label{tab: performance_full}
\resizebox{\textwidth}{!}{%
\begin{tabular}{l|c|cccc?l|c|cccc}
\topline
\textbf{Method} &
\textbf{Backbone} &
  \textbf{RA-PGD} (\%) &
  \textbf{RA-AA} (\%) &  
  \textbf{SA} (\%) &
  \textbf{GFLOPS}(\#) &
  \textbf{Method} &
  \textbf{Backbone} &
  \textbf{RA-PGD} (\%) &
  \textbf{RA-AA} (\%) &  
  \textbf{SA} (\%) &
  \textbf{GFLOPS} (\#)
\\ 
\midtopline \myrowcolour
\multicolumn{12}{c}{\textbf{\texttt{CIFAR-10}}} \\ \midtopline
  {\fullcirc}{\denseRT}  
  & \multirow{5}{*}{ResNet-18} 
  & 50.13\footnotesize{$\pm 0.13$}
  & 44.72\footnotesize{$\pm 0.15$}
  & 82.99\footnotesize{$\pm 0.11$}
  & 0.54
  
  & {\fullcirc}{\denseRT}  
  & \multirow{5}{*}{WRN-28-10} 
  & 51.75\footnotesize{$\pm 0.12$}
  & 45.13\footnotesize{$\pm 0.12$}
  & 83.54\footnotesize{$\pm 0.15$}
  & 5.25 
\\ 
  {\partcirc}{\slimRT} 
  & 
  & 48.12\footnotesize{$\pm 0.09$}
  & 42.24\footnotesize{$\pm 0.13$}
  & 80.18\footnotesize{$\pm 0.11$}
  & 0.14 (74\%$\downarrow$)
  
  & {\partcirc}{\slimRT} 
  & 
  & 50.66\footnotesize{$\pm 0.13$}
  & 44.14\footnotesize{$\pm 0.10$}
  & 82.24\footnotesize{$\pm 0.10$}
  & 1.31 (75\%$\downarrow$)
  
\\
  {\partcirc}{\sparseRT} 
  &
  & 47.93\footnotesize{$\pm 0.17$}
  & 42.11\footnotesize{$\pm 0.11$}
  & \textbf{80.45}\footnotesize{$\pm 0.13$}
  & 0.14 (74\%$\downarrow$)
  
  & {\partcirc}{\sparseRT} 
  & 
  & 48.95\footnotesize{$\pm 0.14$}
  & 43.97\footnotesize{$\pm 0.11$}
  & 82.44\footnotesize{$\pm 0.17$}
  & 1.31 (75\%$\downarrow$)
\\
  {\partcirc}{\moeRT} 
  & 
  & 45.57\footnotesize{$\pm 0.51$}
  & 40.42\footnotesize{$\pm 0.19$}
  & 78.84\footnotesize{$\pm 0.75$}
  & 0.15 (72\%$\downarrow$)
  & {\partcirc}{\moeRT} 
  &
  & 46.73\footnotesize{$\pm 0.46$}
  & 41.11\footnotesize{$\pm 0.23$}
  & 77.42\footnotesize{$\pm 0.73$}
  & 1.75 (67\%$\downarrow$)
\\
\gr {\partcirc}\textbf{{\ours}} 
  &
  & \colorbox{green!30}{\textbf{51.83}}\footnotesize{$\pm 0.12$}
  & \colorbox{green!30}{\textbf{45.13}}\footnotesize{$\pm 0.07$}
  & 80.15\footnotesize{$\pm 0.11$}
  & 0.15 (72\%$\downarrow$)
  
  & {\partcirc}\textbf{{\ours}} 
  &
  & \colorbox{green!30}{\textbf{55.73}}\footnotesize{$\pm 0.13$}
  & \colorbox{green!30}{\textbf{45.89}}\footnotesize{$\pm 0.11$}
  & \colorbox{green!30}{\textbf{84.32}}\footnotesize{$\pm 0.18$}
  & 1.75 (67\%$\downarrow$)
\\
\midtopline \myrowcolour
\multicolumn{12}{c}{\textbf{\texttt{CIFAR-100}}} \\ \midtopline
  {\fullcirc}{\denseRT}  
  & \multirow{5}{*}{ResNet-18} 
  & 27.23\footnotesize{$\pm 0.08$}
  & 23.11\footnotesize{$\pm 0.06$}
  & 58.21\footnotesize{$\pm 0.12$}
  & 0.54
  
  & {\fullcirc}{\denseRT}  
  & \multirow{5}{*}{WRN-28-10} 
  & 27.90\footnotesize{$\pm 0.13$}
  & 23.45\footnotesize{$\pm 0.11$}
  & 57.60\footnotesize{$\pm 0.09$}
  & 5.25 
\\
  {\partcirc}{\slimRT} 
  &
  & 26.41\footnotesize{$\pm 0.16$}
  & 22.11\footnotesize{$\pm 0.13$}
  & 57.02\footnotesize{$\pm 0.14$}
  & 0.14 (74\%$\downarrow$)
  
  & {\partcirc}{\slimRT} 
  &
  & 26.30\footnotesize{$\pm 0.10$}
  & 22.23\footnotesize{$\pm 0.13$}
  & 56.80\footnotesize{$\pm 0.08$}
  & 1.31 (75\%$\downarrow$)
  
\\
  {\partcirc}{\sparseRT} 
  &
  & 26.13\footnotesize{$\pm 0.14$}
  & 21.89\footnotesize{$\pm 0.11$}
  & 57.24\footnotesize{$\pm 0.12$}
  & 0.14 (74\%$\downarrow$)
  
  & {\partcirc}{\sparseRT} 
  & 
  & 25.83\footnotesize{$\pm 0.16$}
  & 21.97\footnotesize{$\pm 0.09$}  
  & 57.39\footnotesize{$\pm 0.14$}
  & 1.31 (75\%$\downarrow$)
\\
  {\partcirc}{\moeRT} 
  &
  & 22.72\footnotesize{$\pm 0.42$}
  & 16.33\footnotesize{$\pm 0.25$}
  & 53.34\footnotesize{$\pm 0.61$}
  & 0.15 (72\%$\downarrow$)
  & {\partcirc}{\moeRT} 
  & 
  & 22.94\footnotesize{$\pm 0.55$}
  & 17.87\footnotesize{$\pm 0.24$}
  & 53.39\footnotesize{$\pm 0.49$}
  & 1.75 (67\%$\downarrow$)
\\
\gr {\partcirc}\textbf{{\ours}} 
  &
  & \colorbox{green!30}{\textbf{28.05}}\footnotesize{$\pm 0.13$}
  & \colorbox{green!30}{\textbf{23.33}}\footnotesize{$\pm 0.06$}
  & \textbf{57.73}\footnotesize{$\pm 0.11$}
  & 0.15 (72\%$\downarrow$)
  
  & {\partcirc}\textbf{{\ours}} 
  &
  & \colorbox{green!30}{\textbf{28.82}}\footnotesize{$\pm 0.14$}
  & \colorbox{green!30}{\textbf{23.57}}\footnotesize{$\pm 0.12$}
  & \textbf{57.56}\footnotesize{$\pm 0.17$}
  & 1.75 (67\%$\downarrow$)
\\ 
\midtopline
\bottomline
\end{tabular}%
}
\vspace*{-4mm}
\end{table*}

\paragraph{Robust evaluation on AutoAttack \cite{croce2020reliable}.} In Tab.\,\ref{tab: performance_full}, we provide additional experiments evaluated by AutoAttack\,\cite{croce2020reliable} (termed \textbf{RA-AA}), a popular robustness evaluation benchmark\,\cite{croce2020robustbench}. The experiment setting in Tab.\,\ref{tab: performance_full} follows \textbf{Tab. \textcolor{red}{1}}. We report RA-AA on {\cifarten} and {\cifarhun} with ResNet-18 and WRN-28-10. As we can see, although AutoAttack leads to a   lower RA-AA compared to RA evaluated using PGD attacks (termed RA-PGD), {\ours} still outperforms {\slimRT}, {\sparseRT}, and {\moeRT} consistently, evidenced by the \textbf{bold} numbers in the  RA-AA columns. 

\begin{figure}[htb]
\vspace*{-3mm}
    \centering
    \begin{tabular}{cc}
    \hspace*{-4mm}\includegraphics[width=.25\textwidth,height=!]{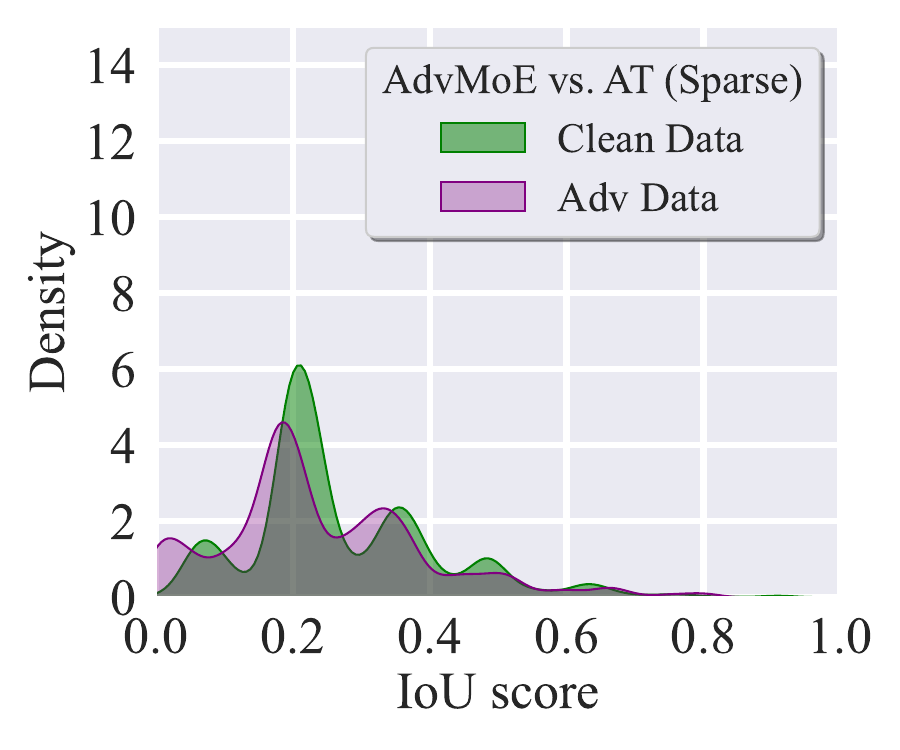}  
    &\hspace*{-4mm}
    \includegraphics[width=.25\textwidth,height=!]{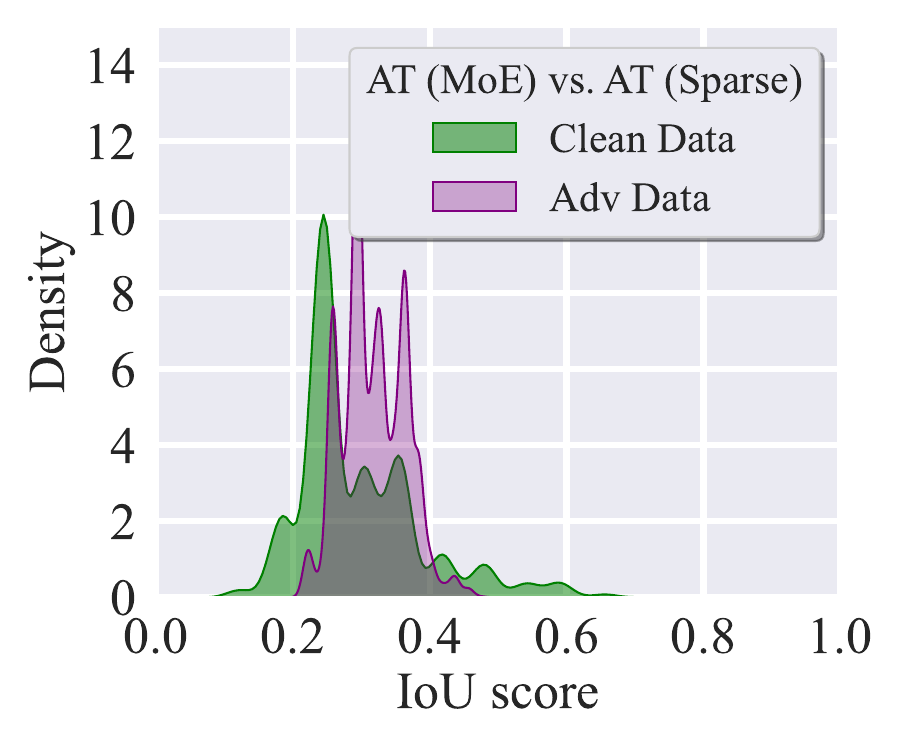} \vspace*{-2mm}\\
    \hspace*{-4mm}\footnotesize{(a) {\ours}} &   \footnotesize{(b) {\moeRT}}
    \end{tabular}
    \caption{{The distribution of the intersection of union (IoU) scores of the input-specific pathways generated by {\ours} (a) and {\moeRT} (b) vs. the static mask found by {\sparseRT}. The distribution over the clean test set and the adversarial test set is plotted for {\moeRT} and {\ours} on setting (ResNet-18, {\cifarhun}). Other settings are aligned with Tab.\ref{tab: exp_main}.}}
    \label{fig: pathway_mask_sim}
    \vspace*{-9mm}
\end{figure}

\paragraph{{\moe} trained by {\ours} enjoys better router utility.}
Based on the results above and the preliminary studies in Sec.\,\ref{sec: methods}, we next peer into the performance difference achieved by {\sparseRT}, {\moeRT}, and {\ours} from the perspective of pathway diversities. We ask: 

\ding{172} What is the relationship between the dynamic pathways generated by the routers trained by {\ours} and the static mask optimized by {\sparseRT}? \ding{173} What is the difference between the routing decisions using {\ours} and   {\moeRT}, and how does it impact the performance? 

Regarding \ding{172}, we investigate the cosine similarity between the pathways generated by training methods, either {{\moeRT}} or {\ours}, and the static mask found by {\sparseRT}. Since the latter can be regarded as a single pathway used for all the data, we term it `\textit{mask pathway}' in contrast to `\textit{MoE pathway}'. We calculate the intersection of union (\textbf{IoU}) score between the MoE pathway and the mask pathway under each testing dataset (the clean or adversarial version).   \textbf{Fig.\,\ref{fig: pathway_mask_sim}} presents the IoU distributions based on the clean and adversarial test datasets (\textbf{Fig.\,\ref{fig: pathway_mask_sim}a} for {\ours} and \textbf{Fig.\,\ref{fig: pathway_mask_sim}b} for {\moeRT}). 
We remark that a smaller IoU score indicates a larger discrepancy between the MoE pathway and the mask pathway. As we can see, the IoU distribution of {\ours}
vs. {\sparseRT} in {Fig.\,\ref{fig: pathway_mask_sim}a} shifts closer to $0$ compared with 
{Fig.\,\ref{fig: pathway_mask_sim}b}. This observation applies to both standard and adversarial evaluation and suggests that {\ours} (our proposal) has a better capability than {\moeRT} to re-build input-specific MoE pathways, which are more significantly different from the input-agnostic mask pathway identified by the pruning-based method, {\sparseRT}. 

Regarding \ding{173}, we observe from Fig.\,\ref{fig: pathway_mask_sim}  that the routers learned by {\moeRT} are more fragile to adversarial attacks compared to {\ours}, as evidenced by the less intersection area of  adversarial data vs. clean data. This   is also aligned with \textbf{Insight 3} in Sec.\ref{sec: methods}.
Moreover, the routing policy learned by  {\ours} is more diverse than {\moeRT}, as indicated by the latter's    density-concentrated IoU scores. In contrast, the distribution of {\ours} is dispersed with a smaller peak value. Therefore, regarding the expert utility, {\ours} is able to assign the inputs to a larger group of pathways than {\moeRT}, making better use of experts.

\paragraph{A coupling effect of expert number $N$ and per-expert model scale $r$ on  {\ours}.}
Recall that there exist two key parameters involved in {\moe} (\textbf{Fig.\,\ref{fig: moe_structure}}): \textit{(a)} the number of experts $N$,  and \textit{(b)} the model scale $r$ that defines the  per-expert (or per-pathway) model capacity. Given  the backbone model (\textit{e.g.}, ResNet-18 in this experiment), a larger $N$  paired with a small $r$   implies that each expert may only have limited model capacity, \textit{i.e.}, corresponding to a less number of channels. Regardless of $N$, if $r = 1$,  the full backbone network will be used to form the identical decision pathway. 

\begin{figure}[htb]
    \centering
    \includegraphics[width=.5\linewidth]{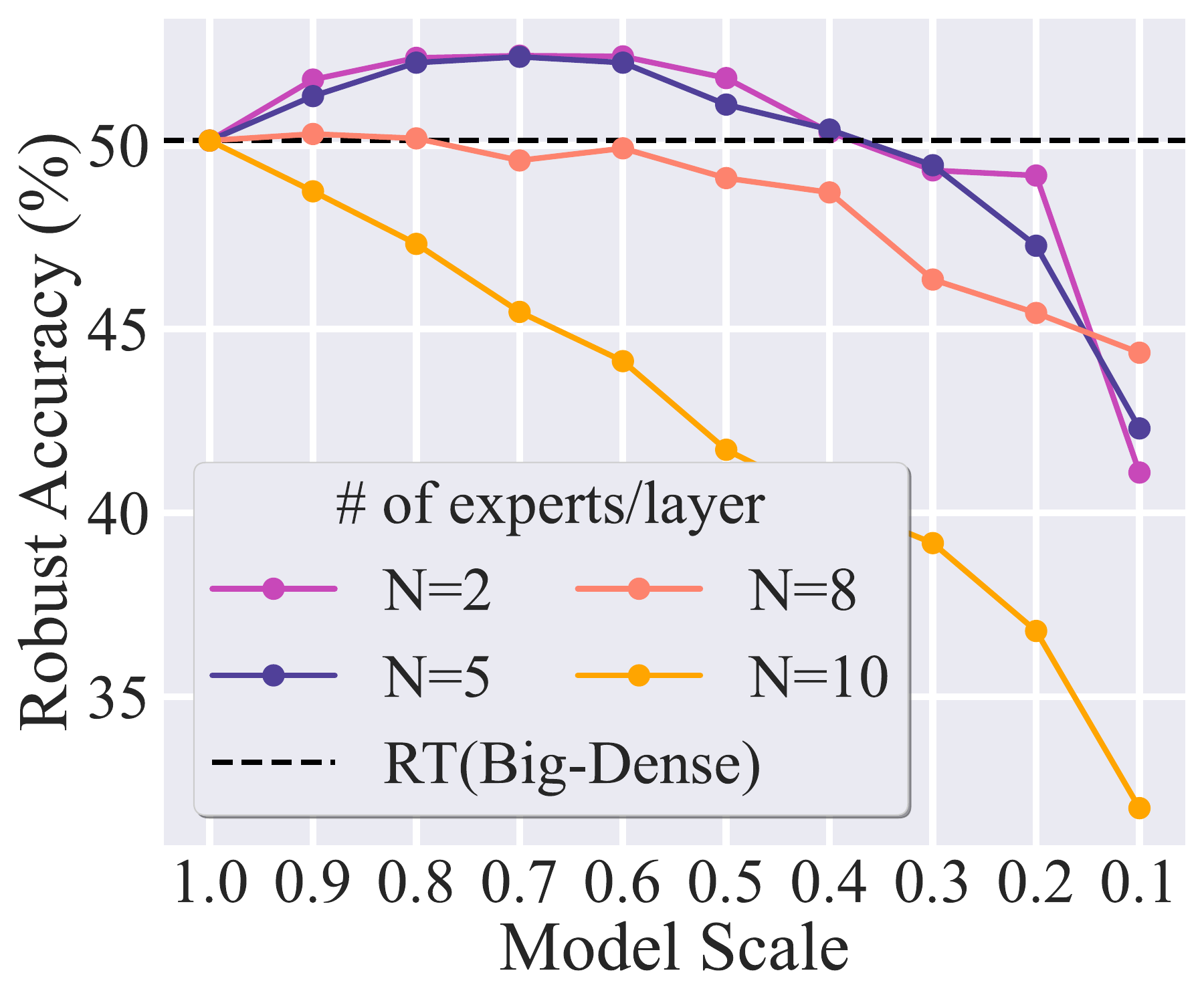}
    \caption{{Performance of {\ours} under {\cifarten} using ResNet-18 as the backbone network for different values of expert number $N$ and model scale $r$. The black dash line denotes the performance of {\ldense} (\textit{i.e.} $r=1$).
    }}
    \label{fig: expert_bank_size}
    \vspace*{-2mm}
\end{figure}

\textbf{Fig.\,\ref{fig: expert_bank_size}} shows the  RA of {\moe} trained by {\ours} vs. the model scale parameter $r$ at different values of $N$. Two insightful observations can be drawn. \textbf{First}, there exists an MoE regime (\textit{e.g.}, $N < 8$ and $r \in [0.5, 0.9]$), in which {\ours} can outperform  {\denseRT} (\textit{i.e.}, $r = 1$) by a substantial margin.
This shows the benefit of MoE in adversarial robustness. However, if the number of experts  becomes larger (\textit{e.g.}, $N = 10$), the increasing diversity of MoE pathways can raise the difficulty of routers' robustification and thus hampers the performance of {\ours} (see $N = 10$ and $r = 0.8$ in \textbf{Fig.\,\ref{fig: expert_bank_size}}). 
\textbf{Second}, there exists an \textit{ineffective} MoE regime (\textit{e.g.}, $N \geq 8$ and $r < 0.5$), in which the performance of {\ours} largely deviates from that of {\denseRT}. In this regime, each expert consists only of a small number of channels, which restricts its robust training ability. Accordingly, both the increasing diversity of MoE pathways  (large $N$) and the limited capacity per pathway (small $r$) could impose    the difficulties of {\AdvRT} for {\moe}. In our experiments, we  choose $r = 0.5$ and $N = 2$, which   preserves the diversity of MoE pathways (\textit{i.e.}, inference efficiency) and retains the effectiveness of robust training.

\begin{figure}[t]
    \centering
    \includegraphics[width=\linewidth]{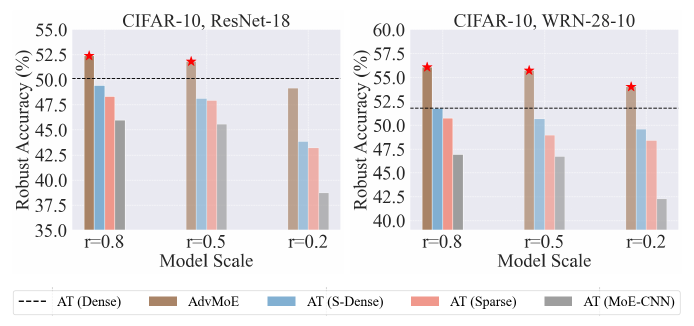}
    \caption{{{Robustness comparison of models trained with different methods under various model scale settings. Results higher than that of {\denseRT} are marked with {\large $\textcolor{red}{\star}$}. Other setups are aligned with Tab.\,\ref{tab: exp_main}. Please refer to Appendix\,\ref{app:more_exps} for exact numbers and GFLOPS comparisons.
    }}}
    \label{fig: exp_scale_study}
    \vspace*{-4mm}
\end{figure}

\paragraph{Performance with different model scales.} To make sure the observations and conclusions from Tab.\,\ref{tab: exp_main} are consistent across different values of the model scale parameter $r$, we repeated the experiments on  ({\cifarten}, ResNet-18) and ({\cifarten}, WRN-28-10) using $r \in \{0.2, 0.5, 0.8\}$ to cover the \{sparse, medium, dense\} regimes with respect to {\ldense} ($r = 1.0$).
\textbf{Fig.\,\ref{fig: exp_scale_study}} summarizes the obtained experiment results. As we can see, {\ours} yields consistent robustness improvements  over all the baselines, including {\ldense}. And the improvement rises as the model scale $r$ increases. This is not surprising as  more parameters will be used when processing one input. Yet, a clear drawback brought by the larger model scale $r$ is the increase of inference cost, evidenced by the  GFLOPS numbers. When $r$ turns to be large (like $r=0.8$), the efficiency benefit brought by the pathway sparsification from   MoE   gradually vanishes. Thus, a medium sparsity ($r=0.5$) is a better choice to balance the trade-off between performance and efficiency, which is thus adopted as our default setting.

\begin{table}[t]
\centering
\caption{{{Performance on robust training for MoE-ViT with in the setup (ImageNet, DeiT-Tiny). Other settings follow Tab.\,\ref{tab: exp_main}.}}}
\label{tab: exp_vit}
\resizebox{.7\linewidth}{!}{%
\begin{tabular}{c|ccc}
\toprule[1pt]
\midrule
{Method} & {RA} (\%) & {SA} (\%) & {GFLOPS} (\#) \\
\midrule
SOTA\cite{puigcerver2022adversarial} & 44.63  & 61.72 & 0.27 \\ 
\gr {\ours} & \textbf{45.93} & 61.67 & 0.27\\
\midrule
\bottomrule[1pt]
\end{tabular}%
}
\vspace*{-5mm}
\end{table}

\paragraph{Extended study:  {\ours} for ViT.} To explore the capability of our proposal {\ours} on ViT-based MoE models (MoE-ViT), \textbf{Tab.\,\ref{tab: exp_vit} } presents   additional results  following the recently published SOTA baseline \cite{puigcerver2022adversarial} for MoE-ViT. As we can see, {\ours} is also applicable to MoE-ViT and can boost  robustness   over the SOTA baseline by over $1\%$ RA improvement, while achieving a similar level of SA. Thus, although our work focuses on robust training for {\moe}, it has the promise of   algorithmic generality to  other {MoE}-based architectures. We defer a more comprehensive study in the future.

\paragraph{Additional experiments.} 
{We conduct ablation studies on (1) robustness evaluation using AutoAttack\,\cite{croce2020reliable} (consistent findings can be drawn as PGD attacks),  (2) attacks steps used in AT, and (3) additional explorations towards the coupling effect between the number of experts and the model scales.  We refer readers to   Appendix\,\ref{app:more_exps} for detailed results.}

\section{Conclusion}
\label{sec: conclusion}
In this work, we design an effective robust training scheme for {\moe}. We first present several key insights on the defense mechanism of {\moe} by dissecting adversarial robustness through the lens of routers and pathways. We next propose {\ours}, the first robust training framework for {\moe} via bi-level optimization, robustifying routers and pathways in a cooperative and adaptive mode. Finally, extensive experiments demonstrate the effectiveness of {\ours} in a variety of data-model setups. Meanwhile, we admit that the {\ours} requires roughly twice the computational capacity compared to the vanilla AT baseline due to alternating optimization that calls two back-propagations per step. Addressing this efficiency concern presents a meaningful avenue for future work.

\section*{Acknowledgement}

The work of Y. Zhang, S. Chang and S. Liu was partially supported by National Science Foundation (NSF) Grant IIS-2207052 and Cisco Research Award. The work of Z. Wang is in part supported by the US Army Research Office Young Investigator Award
(W911NF2010240).

\clearpage\newpage

{\small
\bibliographystyle{unsrt}
\bibliography{ref}
}

\clearpage\newpage
\onecolumn
\section*{\Large{Appendix}}
\setcounter{section}{0}
\setcounter{figure}{0}
\setcounter{table}{0}
\makeatletter 
\renewcommand{\thesection}{\Alph{section}}
\renewcommand{\theHsection}{\Alph{section}}
\renewcommand{\thefigure}{A\arabic{figure}} 
\renewcommand{\theHfigure}{A\arabic{figure}} 
\renewcommand{\thetable}{A\arabic{table}}
\renewcommand{\theHtable}{A\arabic{table}}
\makeatother

\renewcommand{\thetable}{A\arabic{table}}
\setcounter{mylemma}{0}
\renewcommand{\themylemma}{A\arabic{mylemma}}
\setcounter{algorithm}{0}
\renewcommand{\thealgorithm}{A\arabic{algorithm}}
\setcounter{equation}{0}
\renewcommand{\theequation}{A\arabic{equation}}

\section{Sparse MoE Structures}
\label{app: moe_structure}
\begin{figure}[h]
    \centering
    \includegraphics[width=.4\linewidth]{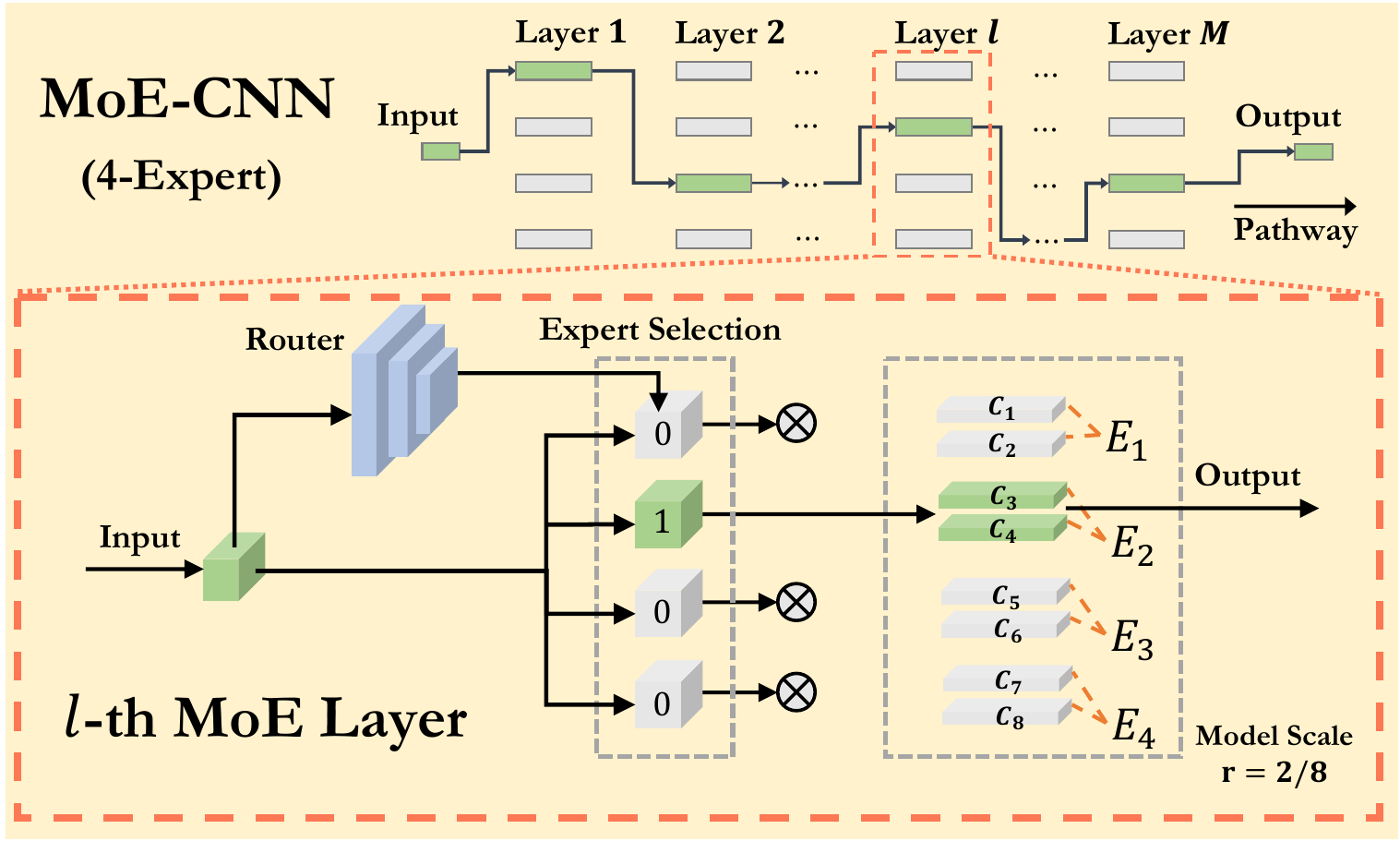}
    \caption{\footnotesize{The sparse {\moe} structure and its MoE design in this paper. The router makes the input-specific expert selection and the selected experts (\textit{e.g.}, $E_2$) form an end-to-end pathway (emphasized in green). This shows an example of the MoE layer with $4$ experts with the model scale of $0.25$.
    }}
    \label{fig: moe_structure}
\end{figure}

\paragraph{Overall MoE Design}
Fig.\,\ref{fig: moe_structure} shows the overall MoE design adopted in this work. By default, the experts in each layer are pre-defined with little channel overlapping. The router exactly selects  one expert for a given input as its   pathway component in this layer.

\section{Detailed Experiment Setups}\label{app:training}

\paragraph{Training Details}
We train all the methods for 100 epochs with an initial learning rate of $0.1$ and a cosine decaying learning rate scheduler. In particular, following the original training pipeline of {\sparseRT} \cite{sehwag2020hydra}, we first train   100 epochs to optimize the mask for {\pcnn}, and then   finetune the model weights based on the fixed mask for another 100 epochs. We use the SGD optimizer for all the methods and a momentum value of $0.9$ together with a weight decay factor of $5e^{-4}$. We use  a batch size of 128  on all the datasets, except 512 for {\imagenet}.

For {\ours}, we randomly sample \textit{different} batches of data (of the same batch size $b$) for updating backbone networks (experts) and routers since the use of diverse data batches is confirmed to benefit generalization for   bi-level learning like     meta-learning\,\cite{deng2021meta} and model pruning\,\cite{zhang2022advancing}.

\paragraph{Datasets and Model Backbones}

To implement {\moe} and other baselines, we conduct experiments on ResNet-18\,\cite{he2016deep}, Wide-ResNet-28-10 (WRN-28-10)\,\cite{zagoruyko2016wide}, VGG-16\,\cite{simonyan2014very}, and DenseNet\,\cite{huang2017densely}.  In particular,   we adopt the ResNet-18 and WRN-28-10 with convolutional kernels of $3 \times 3$ in the first layer for {\timagenet}, {\cifarten} and {\cifarhun}, and $7 \times 7$ for {\imagenet}, following the implementations 
in \cite{resnet2020pytorch}.

\section{Additional Experiments}
\label{app:more_exps}

\paragraph{Ablation study on train-time attack generation steps}
In \textbf{Tab.\,\textcolor{red}{1.}}, we adopt the 2-step PGD attacks to generate the train-time perturbation. Also, we conduct ablation studies on the train-time attack steps and raise its number from $2$ to $10$. We show the obtained results in \textbf{Tab.\,\ref{tab: attack_step}}.
As we can see, the effectiveness of {\ours} holds: Both RA and SA achieved by {\ours} outperform its baselines by a substantial margin.

\begin{table*}[htb]
\centering

\definecolor{rulecolor}{RGB}{0,71,171}
\definecolor{tableheadcolor}{RGB}{204,229,255}
\newcommand{\myrowcolour}{\rowcolor{tableheadcolor}}
\newcommand{\highest}[1]{\textcolor{blue}{\textbf{#1}}}
\newcommand{\topline}{ %
\arrayrulecolor{rulecolor}\specialrule{0.1em}{\abovetopsep}{0pt}%
\arrayrulecolor{rulecolor}\specialrule{\lightrulewidth}{0pt}{0pt}%
\arrayrulecolor{tableheadcolor}\specialrule{\aboverulesep}{0pt}{0pt}%
\arrayrulecolor{rulecolor}}
\newcommand{\midtopline}{ %
\arrayrulecolor{tableheadcolor}\specialrule{\aboverulesep}{0pt}{0pt}%
\arrayrulecolor{rulecolor}\specialrule{\lightrulewidth}{0pt}{0pt}%
\arrayrulecolor{white}\specialrule{\aboverulesep}{0pt}{0pt}%
\arrayrulecolor{rulecolor}}
\newcommand{\bottomline}{ %
\arrayrulecolor{tableheadcolor}\specialrule{\aboverulesep}{0pt}{0pt}%
\arrayrulecolor{rulecolor} %
\specialrule{\heavyrulewidth}{0pt}{\belowbottomsep}     \arrayrulecolor{rulecolor}\specialrule{\lightrulewidth}{0pt}{0pt}%
        }%
\newcolumntype{?}{!{\vrule width 1.5pt}}
\caption{\footnotesize{Ablation study on the train-time attack step numbers. The attack step number used to generate train-time perturbation is raised to $10$ from $2$ compared the default setting. Other settings strictly follow \textbf{Tab\,\textcolor{red}{1}}.
}}
\label{tab: attack_step}

\resizebox{.8\linewidth}{!}{%
\begin{tabular}{l|cccc?cccc}
\topline

\multirow{2}{*}{\textbf{Method}} 
& \multicolumn{4}{c?}{\textbf{ResNet-18} }
& \multicolumn{4}{c}{\textbf{WRN-28-10} }
\\ \cmidrule{2-9}

& \textbf{RA}(\%)  & \textbf{RA-AA}(\%)  & \textbf{SA}(\%) & \textbf{GFLOPS} 
& \textbf{RA}(\%)  &  \textbf{RA-AA}(\%)  & \textbf{SA}(\%) & \textbf{GFLOPS} 
\\ \midtopline 

\myrowcolour
\multicolumn{9}{c}{\textbf{{\cifarten}}} \\ \midtopline

{\fullcirc} \denseRT       
& {50.97}\footnotesize{$\pm0.14$} 
& {46.29}\footnotesize{$\pm0.15$}
& {81.44}\footnotesize{$\pm0.15$} 
& 0.54

& {52.35}\footnotesize{$\pm0.18$} 
& {46.49}\footnotesize{$\pm0.11$}
& {81.45}\footnotesize{$\pm0.15$} 
& 5.25 
\\
                           
{\partcirc} \slimRT       
& {48.22}\footnotesize{$\pm0.11$} 
& {43.79}\footnotesize{$\pm0.15$}
& \textbf{79.93}\footnotesize{$\pm0.12$} 
& 0.14 ($74\%\downarrow$)

& {50.92}\footnotesize{$\pm0.18$} 
& {44.69}\footnotesize{$\pm0.19$}
& {80.33}\footnotesize{$\pm0.15$} 
& 1.31 ($75\%\downarrow$)
\\
                           
{\partcirc} {\sparseRT}    
& {48.29}\footnotesize{$\pm0.14$} 
& {43.18}\footnotesize{$\pm0.19$}
& {79.35}\footnotesize{$\pm0.17$} 
& 0.14 ($74\%\downarrow$)

& {48.69}\footnotesize{$\pm0.18$} 
& {44.50}\footnotesize{$\pm0.16$}
& {80.32}\footnotesize{$\pm0.11$} 
& 1.31 ($75\%\downarrow$)
\\
                           
{\partcirc} {\moeRT}      
& {46.79}\footnotesize{$\pm0.49$} 
& {41.13}\footnotesize{$\pm0.29$}
& {78.32}\footnotesize{$\pm0.51$} 
& 0.15 ($72\%\downarrow$)

& {47.24}\footnotesize{$\pm0.57$} 
& {42.39}\footnotesize{$\pm0.26$}
& {76.21}\footnotesize{$\pm0.42$} 
& 1.75 ($67\%\downarrow$)
 \\
                           
{\partcirc} \textbf{{\ours}}    
& \colorbox{green!30}{\textbf{52.22}}\footnotesize{$\pm0.14$} 
& \colorbox{green!30}{\textbf{46.44}}\footnotesize{$\pm0.09$}
& {79.62}\footnotesize{$\pm0.12$}
& 0.15 ($72\%\downarrow$)

& \colorbox{green!30}{\textbf{56.13}}\footnotesize{$\pm0.11$} 
& \colorbox{green!30}{\textbf{46.73}}\footnotesize{$\pm0.08$}
& \colorbox{green!30}{\textbf{82.19}}\footnotesize{$\pm0.14$} 
& 1.75 ($67\%\downarrow$)
\\

\bottomline
\end{tabular}%
}

\end{table*}

\paragraph{Statistics for Fig.\,\textcolor{red}{7}.} In Fig.\,\textcolor{red}{7}, we show the robustness comparison of different models in various model scale settings.
In Tab.\,\ref{tab: model_scale_study}, we disclose the statistics for the plotting Fig.\,\textcolor{red}{7} as well as the GFLOPS for different model scales.

\begin{table*}[htb]
\centering

\definecolor{rulecolor}{RGB}{0,71,171}
\definecolor{tableheadcolor}{RGB}{204,229,255}
\newcommand{\myrowcolour}{\rowcolor{tableheadcolor}}
\newcommand{\highest}[1]{\textcolor{blue}{\textbf{#1}}}

\newcommand{\topline}{
\arrayrulecolor{rulecolor}\specialrule{0.1em}{\abovetopsep}{0pt}
\arrayrulecolor{rulecolor}\specialrule{\lightrulewidth}{0pt}{0pt}
\arrayrulecolor{tableheadcolor}\specialrule{\aboverulesep}{0pt}{0pt}
\arrayrulecolor{rulecolor}}
\newcommand{\midtopline}{
\arrayrulecolor{tableheadcolor}\specialrule{\aboverulesep}{0pt}{0pt}
\arrayrulecolor{rulecolor}\specialrule{\lightrulewidth}{0pt}{0pt}
\arrayrulecolor{white}\specialrule{\aboverulesep}{0pt}{0pt}
\arrayrulecolor{rulecolor}}
\newcommand{\bottomline}{
\arrayrulecolor{tableheadcolor}\specialrule{\aboverulesep}{0pt}{0pt}
\arrayrulecolor{rulecolor}
\specialrule{\heavyrulewidth}{0pt}{\belowbottomsep}
\arrayrulecolor{rulecolor}\specialrule{\lightrulewidth}{0pt}{0pt}
}
\newcolumntype{?}{!{\vrule width 1.5pt}}
\caption{\footnotesize{
Results of {\ours} (our proposal) \textit{vs.} baselines using different model scale settings on the datasets {\cifarten} and {\cifarhun}. The model scale $r \in \{0.2, 0.5, 0.8\}$ is considered. Other settings strictly follow Tab.\,\ref{tab: performance_full}. The statistics in this table are associated with the plots in Fig.\,\textcolor{red}{7}.
}}
\label{tab: model_scale_study}

\resizebox{\linewidth}{!}{
\begin{tabular}{l|ccc|ccc|ccc?ccc}
\topline

\multirow{2}{*}{\textbf{Method}} 
& \multicolumn{3}{c|}{\textbf{model scale $r=0.2$} }
& \multicolumn{3}{c|}{\textbf{model scale $r=0.5$} }
& \multicolumn{3}{c?}{\textbf{model scale $r=0.8$} }
& \multicolumn{3}{c}{\textbf{\denseRT}, model scale $r=1.0$}      
\\ \cmidrule{2-13}

& \textbf{RA}(\%)  & \textbf{SA}(\%) & \textbf{GFLOPS} 
& \textbf{RA}(\%)  & \textbf{SA}(\%) & \textbf{GFLOPS} 
& \textbf{RA}(\%)  & \textbf{SA}(\%) & \textbf{GFLOPS} 
& \textbf{RA}(\%) & \textbf{SA}(\%) & \textbf{GFLOPs} 
\\ \midtopline 

\myrowcolour
\multicolumn{13}{c}{\textbf{{\cifarten}, ResNet-18}} \\ \midtopline
                           
\slimRT       
& 43.83\footnotesize{$\pm 0.11$} 
& 78.28\footnotesize{$\pm 0.14$} 
& 0.13 ($76\%\downarrow$) 

& 48.12\footnotesize{$\pm 0.09$} 
& 80.18\footnotesize{$\pm 0.11$} 
& 0.14 ($74\%\downarrow$)

& 49.44\footnotesize{$\pm 0.09$} 
& 81.32\footnotesize{$\pm 0.11$} 
& 0.36 ($33\%\downarrow$)

& \multirow{4}{*}{50.13\footnotesize{$\pm 0.13$} } 
& \multirow{4}{*}{82.99\footnotesize{$\pm 0.11$} } 
& \multirow{4}{*}{0.54} 
\\
                           
{\sparseRT}    
& 43.24\footnotesize{$\pm 0.14$} 
& \textbf{79.14}\footnotesize{$\pm 0.14$} 
& 0.13 ($76\%\downarrow$)

& 47.93\footnotesize{$\pm 0.17$} 
& \textbf{80.45}\footnotesize{$\pm 0.13$} 
& 0.14 ($74\%\downarrow$)

& 48.32\footnotesize{$\pm 0.13$} 
& 81.77\footnotesize{$\pm 0.11$} 
& 0.36 ($33\%\downarrow$)
\\
                           
{\moeRT}      
& 38.75\footnotesize{$\pm 0.41$} 
& 76.54\footnotesize{$\pm 0.29$} 
& 0.14 ($74\%\downarrow$)

& 45.57\footnotesize{$\pm 0.51$} 
& 78.84\footnotesize{$\pm 0.75$} 
& 0.15 ($72\%\downarrow$)

& 45.99\footnotesize{$\pm 0.42$} 
& 79.46\footnotesize{$\pm 0.31$} 
& 0.37 ($31\%\downarrow$)
\\
                           
\textbf{{\ours}}    
& \textbf{49.18}\footnotesize{$\pm 0.12$} 
& \textbf{79.03}\footnotesize{$\pm 0.19$} 
& 0.14 ($74\%\downarrow$)

& \colorbox{green!30}{\textbf{51.83}}\footnotesize{$\pm 0.12$} 
& 80.15\footnotesize{$\pm 0.11$} 
& 0.15 ($72\%\downarrow$) 

& \colorbox{green!30}{\textbf{52.38}}\footnotesize{$\pm 0.14$} 
& 81.44\footnotesize{$\pm 0.13$} 
& 0.37 ($31\%\downarrow$)
\\ \midtopline 

\myrowcolour
\multicolumn{13}{c}{\textbf{{\cifarten}, WRN-28-10}} 
\\ \midtopline 
                           
\slimRT 
& 49.59\footnotesize{$\pm 0.17$} 
& \textbf{79.93}\footnotesize{$\pm 0.13$} 
& 0.21 ($96\%\downarrow$)

& 50.66\footnotesize{$\pm 0.13$} 
& 82.24\footnotesize{$\pm 0.10$} 
& 1.31 ($75\%\downarrow$)

& 51.73\footnotesize{$\pm 0.17$} 
& 82.88\footnotesize{$\pm 0.14$} 
& 3.36 ($38\%\downarrow$)

& \multirow{4}{*}{51.75\footnotesize{$\pm 0.12$}} 
& \multirow{4}{*}{83.54\footnotesize{$\pm 0.15$}}
& \multirow{4}{*}{5.25} 
\\
                           
{\sparseRT} 
& 48.37\footnotesize{$\pm 0.21$} 
& 79.32\footnotesize{$\pm 0.21$} 
& 0.21 ($96\%\downarrow$)

& 48.95\footnotesize{$\pm 0.14$} 
& 82.44\footnotesize{$\pm 0.17$} 
& 1.31 ($75\%\downarrow$)

& 50.73\footnotesize{$\pm 0.19$} 
& 82.11\footnotesize{$\pm 0.23$} 
& 3.36 ($38\%\downarrow$)
\\

{\moeRT}  
& 42.29\footnotesize{$\pm 0.51$} 
& 75.32\footnotesize{$\pm 0.38$} 
& 0.94 ($82\%\downarrow$)

& 46.73\footnotesize{$\pm 0.46$} 
& 77.42\footnotesize{$\pm 0.73$} 
& 1.75 ($67\%\downarrow$)

& 46.94\footnotesize{$\pm 0.45$} 
& 79.11\footnotesize{$\pm 0.27$} 
& 4.57 ($13\%\downarrow$)
\\
                           
\textbf{{\ours}} 
& \textbf{54.02}\footnotesize{$\pm 0.09$} 
& 79.55\footnotesize{$\pm 0.12$} 
& 0.94 ($82\%\downarrow$)

& \colorbox{green!30}{\textbf{55.73}}\footnotesize{$\pm 0.13$}
& \colorbox{green!30}{\textbf{84.32}}\footnotesize{$\pm 0.18$} 
& 1.75 ($67\%\downarrow$)

& \colorbox{green!30}{\textbf{56.07}}\footnotesize{$\pm 0.14$} 
& \colorbox{green!30}{\textbf{84.45}}\footnotesize{$\pm 0.09$} 
& 4.57 ($13\%\downarrow$)
\\ \midtopline
\bottomline
\end{tabular}%
}
\end{table*}

\begin{figure}[h]
    \centering
    \begin{tabular}{cc}
    \hspace*{-4mm}\includegraphics[width=.25\textwidth,height=!]{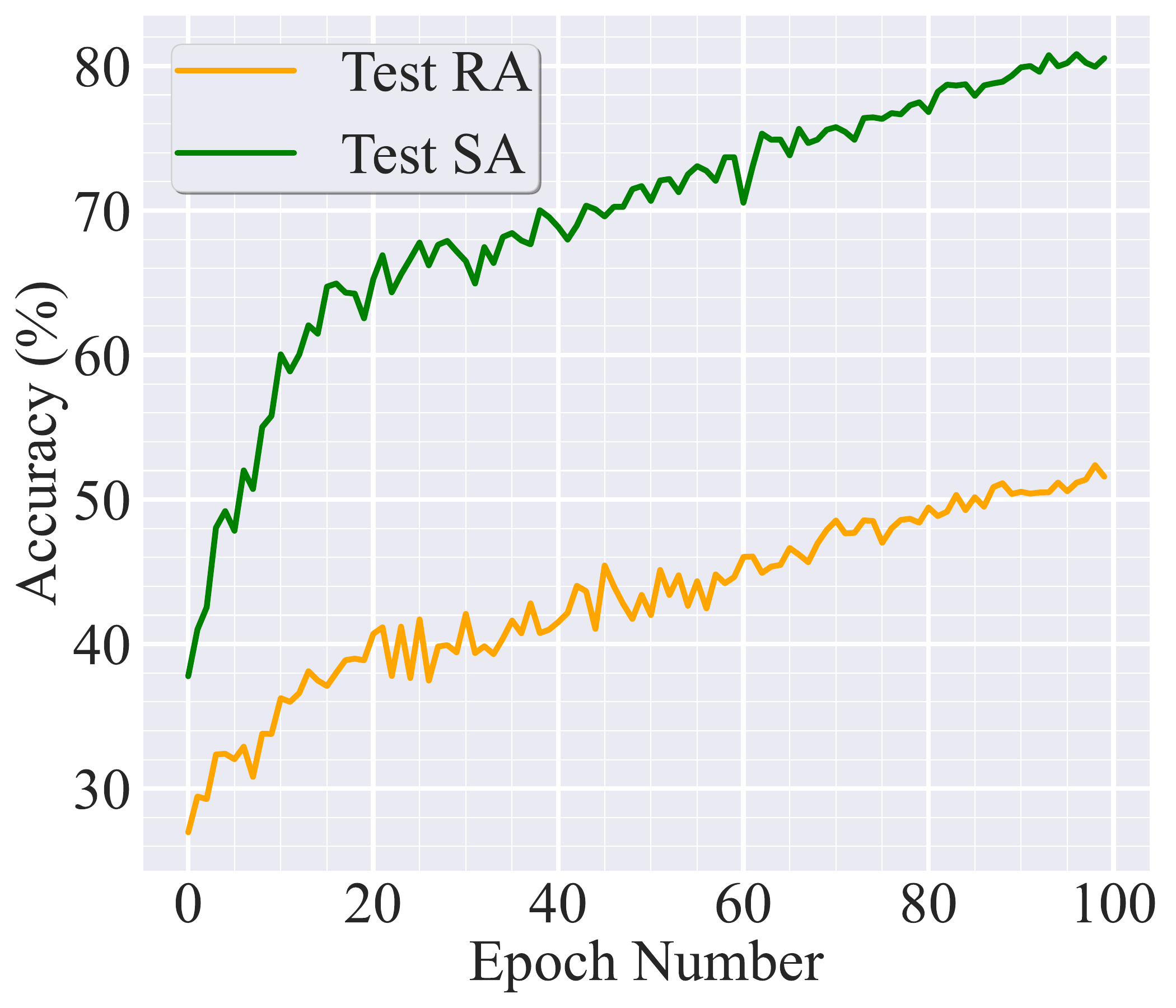}  
    &\hspace*{-4mm}
    \includegraphics[width=.25\textwidth,height=!]{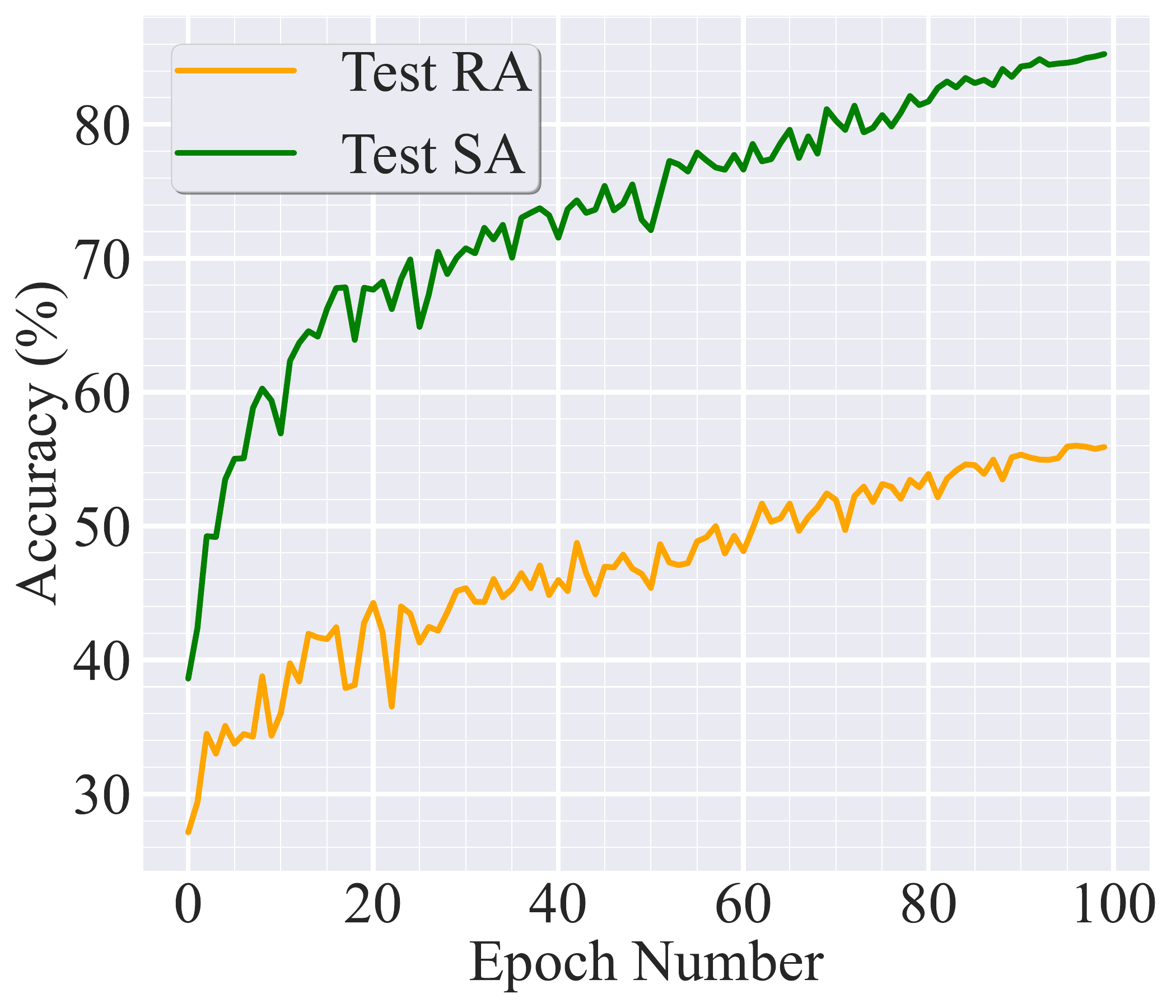} \vspace*{-2mm}\\
    \hspace*{-4mm}\footnotesize{(a) ResNet-18, CIFAR10} &   \footnotesize{(b) WRN-28-10, CIFAR-10}
    \end{tabular}
    \caption{\footnotesize{The training trajectory of {\ours} under different data-model settings. }}
    \label{fig: convergence}
\end{figure}

\paragraph{Training trajectory {\ours}.} We show in Fig.\,\ref{fig: convergence} that the {\ours} converges well within 100 training epochs using a cosine learning rate schedule. The SA (standard accuracy) and RA (robust accuracy) 
are evaluated and collected at the end of each training epoch.

\end{document}